\newcommand{\specialcell}[2][l]{%
  \begin{tabular}[#1]{@{}l@{}}#2\end{tabular}}

\documentclass[phd,tocprelim]{cornell}
\let\ifpdf\relax
\usepackage{graphicx,pstricks}
\usepackage{graphics}
\usepackage{moreverb}
\usepackage{epsfig}
\usepackage{caption}
\usepackage{subfigure}
\usepackage{hangcaption}
\usepackage{palatino}
\usepackage{hyperref}
\usepackage{amsmath}
\usepackage{float}
\usepackage{booktabs} 
\usepackage{multirow}
\usepackage{titlesec}
\usepackage{balance}
\usepackage{fancyhdr}
\usepackage[utf8]{inputenc}

\usepackage[justification=justified]{caption}
\usepackage[authoryear,sort]{natbib}

\renewcommand{\caption}[1]{\singlespacing\hangcaption{#1}\normalspacing}

\title {Towards Understanding Persuasion in Computational Argumentation}
\author {Esin Durmus}
\conferraldate {May}{2021}
\degreefield {Ph.D.}
\copyrightholder{Esin Durmus}
\copyrightyear{2021}

\begin{document}

\maketitle
\makecopyright

\begin{abstract}
Opinion formation and persuasion in argumentation are affected by three major factors: the argument itself, the source of the argument, and the properties of the audience. Understanding the role of each and the interplay between them is crucial for obtaining insights regarding argument interpretation and generation. It is particularly important for building effective argument generation systems that can take both the discourse and the audience characteristics into account. Having such personalized argument generation systems would be helpful to expose individuals to different viewpoints and help them make a more fair and informed decision on an issue.

Even though studies in Social Sciences and Psychology have shown that source and audience effects are essential components of the persuasion process, most research in computational persuasion has focused solely on understanding the characteristics of persuasive language. In this thesis, we make several contributions to understand the relative effect of the source, audience, and language in computational persuasion. We first introduce a large-scale dataset with extensive user information to study these factors' effects simultaneously. Then, we propose models to understand the role of the audience's prior beliefs on their perception of arguments. We also investigate the role of social interactions and engagement in understanding users' success in online debating over time. We find that the users' prior beliefs and social interactions play an essential role in predicting their success in persuasion. Finally, we explore the importance of incorporating contextual information to predict argument impact and show improvements compared to encoding only the text of the arguments. 

\end{abstract}

\begin{biosketch}
Esin was born in Bursa, Turkey. She obtained a Bachelor's degree in Industrial Engineering and Computer Engineering at Koç University, where she was the valedictorian of her class. She was introduced to computers at an early age since her father is a Computer Engineer, and he always motivated her to follow his path. During her undergrad, she explored several research areas such as Computer Verification and Intelligent User Interfaces. However, she was not aware of Natural Language Processing (NLP) research until she started her Ph.D. at Cornell. 

During her Ph.D., she took several courses in Natural Language Processing to better understand the state of research. She was very impressed that NLP provided an opportunity to integrate machine learning methods to understand social phenomena such as persuasion. She spent a lot of time thinking about computationally modeling persuasion during her Ph.D. She has always valued collaborations since she learned a lot from other researchers in this field. She did internships at Amazon and Google, where she met a fantastic group of people and worked on text generation and evaluation which motivated her to do a Postdoc at Stanford University to explore these areas further. Apart from research, Esin enjoys watching movies, traveling, trying food from different cuisines, and playing table tennis.  

\end{biosketch}

\begin{dedication}
To my parents and Faisal for being there through the journey.
\end{dedication}

\begin{acknowledgements}
First and foremost, I am incredibly grateful to my advisor, Claire Cardie, for her support, kindness, patience, and guidance. I was not familiar with Machine Learning (ML) and Natural Language Processing (NLP) until I started my Ph.D. During my first year, I took the "Introduction to Natural Language Processing" course taught by Claire. That course was instrumental in my decision to pursue a career in the field. Claire has always been very patient and understanding. She always gave me the time and opportunity to learn and explore new research directions. I am forever grateful to have had her support during the beginning of my career.  I am also very thankful to my committee members John Hopcroft and Cristian Danescu-Niculescu-Mizil for their feedback on my research.

I was fortunate to do great internships at Amazon and Google during my Ph.D. I want to thank my collaborators and mentors, He He, Mona Diab, Kathleen McKeown, Smaranda Muresan at Amazon AWS, and Gaurav Kumar, Ankur Parikh, Diyi Yang, Sebastian Gehrmann, Dipanjan Das at Google Research, who helped me gain a more diverse perspective about research in NLP. I am incredibly grateful to He He and Mona Diab for supporting and guiding me in my career decisions beyond the internship. I am fortunate to have their mentorship. 

I want to thank Ana Smith, Arzoo Katiyar, Ashudeep Singh, Jialu Li, Kai Sun, Liane Longpre, Maria Antoniak, Ozan Irsoy, Rishi Bommasani, Tianze Shi, Vlad Niculae, Xilun Chen, Xinran Zhao, Xinya Du, Yao Cheng, and all the members of the NLP seminar for fruitful discussions and providing feedback on my research. 

I have been fortunate to be surrounded by amazing friends. Just to name a few: Deniz Altınbüken, Burcu Çanakçı, Connie Ong Blukis, Valts Blukis, Kim Bomin, Milla Vastavuo, Parsifal Islas, Ilse Alejo, Pablo Lujambio, Faisal Alkaabneh, Nouha Dziri, Amr Sharaf, Tuhin Chakrabarty, Yiqing Hua, Andreas Veit, Eugune Bagdasaryan, Nadia Bagdasaryan, Julia Tolkacheva, Ilia Ilmer and Nidhi Baid. I am forever grateful to my friends who have been instrumental in keeping me motivated and pulling me out of stressful moments of life. I would not have been able to do any of this without them.

I want to thank my parents: Eşref Durmuş and Tülay Durmuş. They have always motivated me to pursue a career in Computer Science and given me tremendous support throughout this journey. I can never pay back their support. 

Lastly, I would like to thank Faisal Ladhak for his endless love and support. I have been very fortunate to have him by my side throughout this journey. Through these years, we experienced life in four different cities (Ithaca, Seattle, NYC, and Mountain View), and I enjoyed every moment of my time with him. Besides being my life partner and best friend, he has also been a fantastic collaborator during these years. He always inspires me and brings the best out of me. I am forever grateful to him. 

\end{acknowledgements}

\contentspage
\tablelistpage
\figurelistpage

\normalspacing \setcounter{page}{1} \pagenumbering{arabic}
\pagestyle{cornell} \addtolength{\parskip}{0.5\baselineskip}

\chapter{Introduction}

\section{Background}

Argumentation is a discussion in which reasons are provided for and against some proposition or proposal \citep{Toulmin1958-TOUTUO-2}. It is a crucial activity in decision-making since it encourages critical thinking and motivates people to make fair and informed decisions.
The emergence of social media and online argumentation platforms has made it easier for people to express their opinions and debate with other individuals on controversial topics. The reliance on social media and online argumentation platforms as key venues for opinionated discussions \citep{elisa_2020} has motivated increased research in computational argumentation. 
One area of focus in computational argumentation has been to explore techniques to automatically analyze the characteristics of arguments, such as their persuasiveness \citep{wachsmuth-etal-2017-computational, habernal-gurevych-2016-makes, 10.1145/2872427.2883081, DBLP:conf/argmining/HideyMHMM17}. Furthermore, researchers have started building systems to automatically generate arguments to present people with diverse viewpoints in order to help them make more informed decisions  \citep{wang-ling-2016-neural, hua-wang-2018-neural, hua-etal-2019-argument, hidey-mckeown-2019-fixed}. 

This thesis focuses on understanding the factors of persuasion in computational argumentation. Persuasion is an act of presenting arguments to change people's opinions, values, and behaviors on a controversial topic or an event  \citep{edsjsr.208777219540601}. 
Theories of persuasive communication are applied to various fields such as marketing, advertising, social psychology, and politics \citep{Shrum2012PersuasionIT}. Researchers in these areas are interested in understanding the factors that influence the success of persuasive communication.
The emergence of social media and online argumentation platforms has made it more accessible for people to engage in argumentative discussions with others who may hold differing views.
Interpretation of the underlying dynamics of argumentative communication online can help develop methods to improve the effectiveness of arguments. 
For example, it could be used to provide feedback to users in order to help them improve the structure of their arguments.
Moreover, analyzing these interactions can help understand the factors that influence people's behavior in the argumentative process. We specifically study persuasion on online debating platforms to get insights into the factors that govern people's decision-making in persuasion.

Language is the primary tool that is used to convey the content of an argument. Therefore it is a crucial component in persuasion \citep{1984-28616-00119840101, edsovi.00005205.198111000.0000219811101, perelman1971new,van2009examining}.
It is only natural then that the majority of the work in computational studies of persuasion has focused mainly on understanding the characteristics of persuasive text, e.g., what distinguishes persuasive from non-persuasive text \citep{10.1145/2872427.2883081, zhang2016conversational, wachsmuth2016using, habernal-gurevych-2016-makes,habernal-gurevych-2016-argument,fang2016learning,DBLP:conf/argmining/HideyMHMM17}. However, language is not the only factor in persuading people. Prior research in Social Sciences and Psychology has shown that the recipients of an argument may form their opinion on an issue based on non-content cues such as the characteristics of the speaker (i.e., the source) and their own predispositions \citep{cialdini2001influence, 1984-28616-00119840101, edsovi.00005205.198111000.0000219811101}. For example, the credibility and trustworthiness of a speaker \citep{edsgcl.1760738119951001, edsovi.00005205.198011000.0000119801101} and the prior beliefs of the audience  \citep{edsbds.17414237419960101} have been shown to have a substantial effect on persuasive communication. Furthermore, people with strong prior beliefs on controversial issues have been shown to have biased stances even when presented with empirical evidence: i.e., they tend to find empirical evidence that confirms their prior beliefs more convincing \citep{edsovi.00005205.197911000.0001619791101}. Given the evidence from Social Sciences and Psychology, we believe that accounting for the impact of these factors, in addition to the language, in computational studies of persuasion is crucially important. This thesis introduces several contributions to make progress towards this goal by exploring the following questions:

\begin{itemize}
    \item What is the role of people's prior beliefs and initial stance on persuasion? 
    \item How can we disentangle the effect of source and audience factors in order to understand the characteristics of persuasive language? 
    \item What is the effect of social interaction on people's success at persuasion over time? 
    \item Does pragmatic context play an important role in predicting the impact of arguments? 
\end{itemize}

\section{Contributions}

In order to understand the role of speaker and audience effects in persuasion, we primarily look at the following factors of interaction on online argumentation: 
\begin{enumerate}
    \item Prior beliefs and initial stances of the speaker and the audience.
    \item Social interactions.  
    \item Language and pragmatic context. 
\end{enumerate}

\begin{figure*}[]
\centering
\includegraphics[scale=0.83]{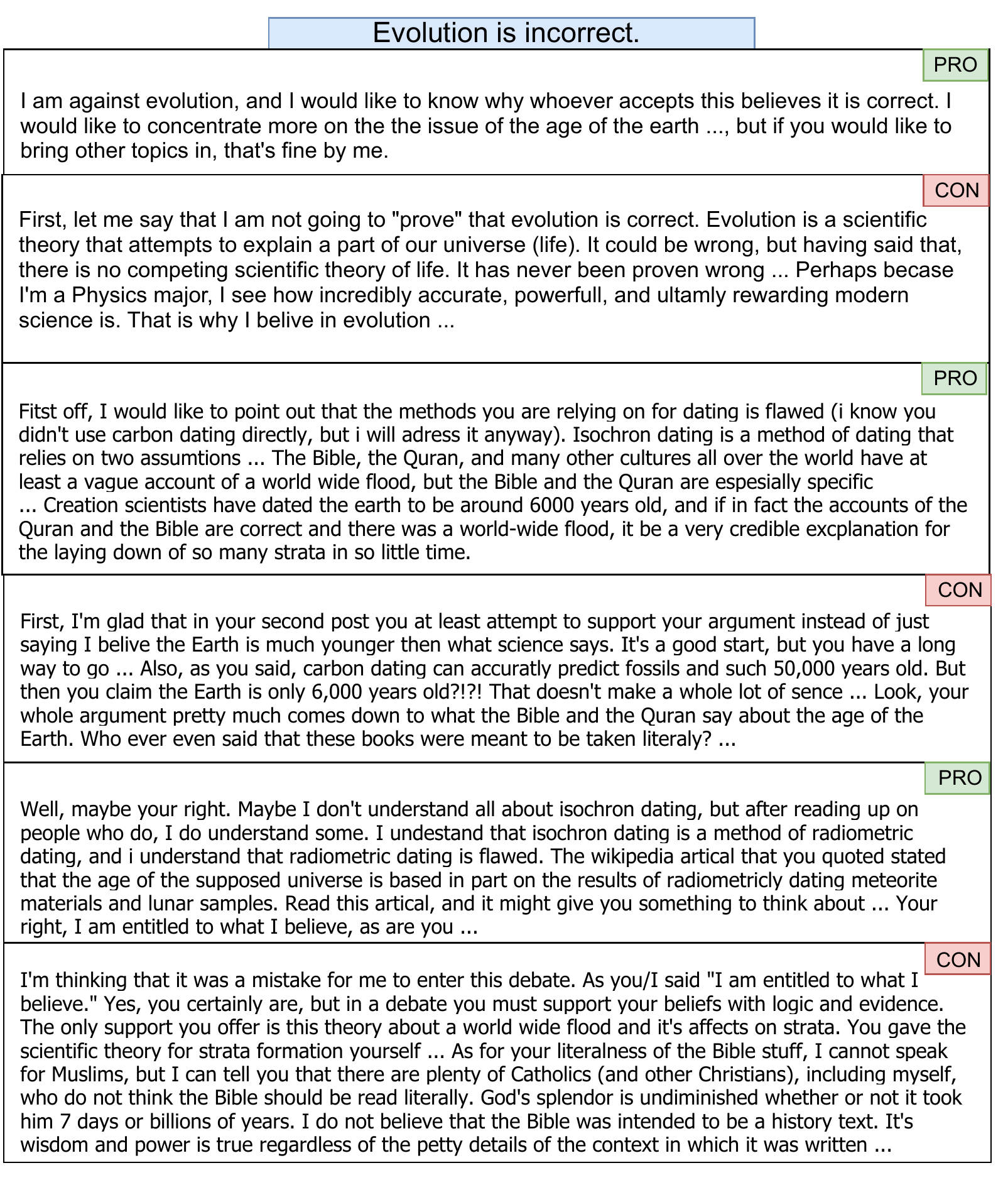}
\caption{A debate on "Evolution".}
\label{fig:example_debate}
\end{figure*}
\newpage
Specifically, we make the following contributions:

\textbf{A Dataset to Model Source and Audience Factors in Persuasion.} One of the main bottlenecks in studying the effect of source and audience factors is the lack of large-scale datasets that contain information about the characteristics of the users. In order to bridge this gap and enable further studies in this area, we present a large-scale dataset (DDO) with a wide range of user information collected from an online debating website.\footnote{\url{https://www.debate.org}} This dataset contains debates on a wide range of controversial topics. Each debate consists of two debaters with opposing views on a controversial topic, who take turns to provide their arguments. Figure \ref{fig:example_debate} shows an example debate on \textit{"Evolution"} contained in this dataset. Along with the debate, the dataset also contains votes from the audience evaluating various aspects of the debaters, such as the persuasiveness of their arguments and their overall debating skills. Besides the debates, the dataset also includes information about the debaters and the audience, such as their stance on controversial topics, political and religious ideologies, education level, etc. We obtain this from the self-identified information that the users provide on their profiles. 

\textbf{The Role of Prior Beliefs in Persuasion.} The majority of work in computational persuasion has focused on understanding the characteristics of persuasive language. In this thesis, we mainly focus on understanding the effect of user factors on persuasion. We use the debates, votes, and user information available on the DDO dataset to study the effect of prior beliefs in predicting which debater an individual voter will find more persuasive for a given debate. We find that user factors play a critical role in this prediction task. Furthermore, controlling for the effect of user-level factors allows us to investigate characteristics of persuasive language without any influence from these potentially confounding factors.

\textbf{Effect of Social Interaction on Persuasion Success.} Inspired by prior work that shows a strong relationship between a user's social interaction and their influence on social media \citep{cha2010measuring,10.1145/1963192.1963250}, we study whether success in persuasion might also depend on an individual's social interaction and engagement. 
In particular, we study whether users can improve the persuasiveness of their arguments as they gain more experience using the debating platform. We show that a user's social interaction is an essential factor in predicting their overall success at debating.

\textbf{Representing Pragmatic Context in Modeling Argument Impact.} We present a new dataset to study the effect of the pragmatic and discourse context
when determining an argument's impact. We further propose predictive models that can incorporate pragmatic and discourse context. We find that these models outperform models that rely only on claim-specific linguistic features for predicting the perceived impact of individual claims within a particular line of argument.

\section{Organization of Thesis}

In Chapter \ref{prior_work}, we first give an overview of recent developments in computational argumentation, describing recent work on argument analysis and argument generation. In Chapter \ref{dataset}, we discuss the details and statistics of the DDO dataset. With the dataset in hand, in Chapter \ref{prior_beliefs}, we present methods that can account for the effects of the speaker and the audience in predicting the persuasiveness of an argument. In Chapter \ref{social_interactions}, we describe our contributions on understanding the impact of social interaction on persuasion on online debating platforms. In Chapter \ref{context}, we propose a new dataset that allows us to study the effect of pragmatic context (i.e., \textit{kairos}) on assessing the impact of an argument. 
Finally, in Chapter \ref{conclusion}, we summarize our contributions and provide directions for future work.

\chapter{An Overview of Computational Argumentation}\label{prior_work}

This chapter describes the recent developments in various sub-fields of computational argumentation, including computational argumentation mining, computational persuasion, and automated argument generation.

\section{Computational Argument Mining}

Computational argumentation mining aims to extract argument components and the relationships between them from unstructured text, building on theoretical models of argument \citep{Toulmin1958-TOUTUO-2,walton_reed_macagno_2008}. The main goal is to understand the points in an argument and get insights into how these points support or oppose each other. Having a deeper understanding of the structure of the arguments is important for various applications such as debating technologies \citep{PMID:33731946_debater}, legal decision-making \citep{10.1145/1276318.1276362}, automated essay scoring \citep{ong-etal-2014-ontology}, and computer-assisted writing \citep{stab-gurevych-2017-parsing}. The identification of argument structure involves several sub-tasks:

\begin{enumerate}
    \item Determining the ``argumentative'' vs. ``non-argumentative'' parts of the text \citep{10.1145/1276318.1276362}. 
    \item Classifying argumentative components into categories such as ``Claim'' or ``Premise'' \citep{article_moens, stab-gurevych-2017-parsing,chakrabarty-etal-2019-imho}.
    \item Identifying relations between argument components \citep{carstens-toni-2015-towards,feng2011classifying, 10.1145/1568234.1568246, Cabrio2013ANL, stab-gurevych-2017-parsing, niculae-etal-2017-argument, park-cardie-2014-identifying,hua-wang-2017-understanding}. 
\end{enumerate}

Most research in computational argumentation mining has proposed methods for a subset of the subtasks mentioned above. \cite{persing-ng-2016-end} was among the first to present an end-to-end pipeline approach to determine argumentative components and their relationship using an Integer Linear Programming (ILP) framework. Similarly, \cite{stab-gurevych-2017-parsing} has proposed a joint model that globally optimizes argument component types and relations using ILP.  \cite{eger-etal-2017-neural} has presented the first end-to-end neural argumentation mining model obviating the need for designing hand-crafted features and constraints. 
 
Argumentation mining has been applied to various domains such as persuasive essays, legal documents, political debates, and social media data \citep{dusmanu-etal-2017-argument}. For instance, \cite{stab-gurevych-2017-parsing} has built an annotated dataset of persuasive essays with corresponding argument components and relations. Using this corpus, \cite{eger-etal-2017-neural} developed an end-to-end neural method for argument structure identification. \cite{DBLP:conf/aaai/NguyenL18} has further proposed an end-to-end method to parse argument structure and used the argument structure features to improve automated persuasive essay scoring. Furthermore, \cite{levy-etal-2014-context} has studied context-dependent claim detection by collecting annotations for Wikipedia articles. Using this corpus, \cite{rinott-etal-2015-show} has investigated the task of automatically identifying the corresponding pieces of evidence given a claim.  \cite{bar-haim-etal-2017-stance} has further proposed the task of claim-stance detection (i.e., given a topic and claims, identifying for each claim whether it supports or opposes the topic.) by further annotating Wikipedia articles with stance information. \cite{walker-etal-2012-corpus} has collected posts from \href{www.4forums.com}{\textit{4forums.com}}, a debating forum, and have further annotated part of this corpus for various aspects of arguments such as topic, stance, agreement, and sarcasm. \cite{park-cardie-2018-corpus} has proposed an argument mining corpus from  Consumer Debt Collection Practices (CDCP) rule by the Consumer Financial Protection Bureau (CFPB) posted on
\href{www.regulationroom.org}{\textit{regulationroom.org}}. Using this corpus, \cite{niculae-etal-2017-argument} proposed a structured prediction model for argumentation mining.   

Although most of the research in Argumentation Mining has focused on English monologues, \cite{Peldszus2015AnAC} has collected a corpus of microtexts in German and used this corpus for argument component detection. Furthermore, \cite{basile:hal-01414698} has studied relation prediction task in Italian news blogs. Similarly, there has been some recent work investigating argumentation mining beyond monologues, i.e., looking at the process of argumentation in dialogues. For example, \cite{chakrabarty-etal-2019-ampersand} has proposed a method to
identify the argument structure in persuasive dialogues that can model the micro-level (i.e., the structure of a single argument) and the macro-level (i.e., the interplay between the arguments) characteristics of arguments. 

Stance detection and Argumentation Mining are closely related tasks, given that they both aim to understand standpoints from the text on a controversial topic. Contrary to stance detection, argumentation mining aims also to extract a more fine-grained structure of arguments, identifying claims, premises, and the relationship between them. There has been a lot of research on identifying the stance of arguments on a controversial topic \citep{sobhani-etal-2015-argumentation,hasan-ng-2013-stance, bar-haim-etal-2017-improving,sun-etal-2018-stance}.  For example, \cite{sobhani-etal-2015-argumentation} has shown that using argument structure features improves the performance of stance detection models. \cite{wachsmuth-etal-2018-retrieval} has further studied retrieval of the best counter-arguments, using arguments opposing the same aspect of the controversial topic. 
In our work \citep{durmus-etal-2019-determining}, we have found that encoding contextual information using the argument structure tree is crucial to achieving state-of-the-art performance for argument stance detection. \cite{kobbe-etal-2020-unsupervised} has proposed an unsupervised method to assess the stance of the arguments inferring whether the outcome is good vs. bad. 

For a more detailed discussion of the argumentation mining literature, refer to comprehensive surveys by \cite{10.4018/jcini.2013010101}, \cite{ijcai2018-766}, and \cite{lawrence-reed-2019-argument}.

\section{Computational Studies of Persuasion}

Understanding the characteristics of persuasive language has been a great interest of computational studies of persuasion. Most of the work in this domain has focused solely on language \citep{DBLP:conf/argmining/HideyMHMM17, habernal-gurevych-2016-makes,guerini-etal-2015-echoes, li-etal-2020-exploring-role, el-baff-etal-2020-analyzing, 10.1145/2872427.2883081, atkinson-etal-2019-gets,morio-etal-2019-revealing, 10.5555/3171837.3171856}. For instance, \cite{habernal-gurevych-2016-argument} has collected a new corpus to study the task of predicting
which argument from an argument pair is the more convincing. \cite{zhang2016conversational} has studied the role of conversational flow and interplay between debaters on persuasion in Oxford-style debates. \cite{DBLP:conf/aaai/HideyM18} has further investigated the role of larger context on persuasion, modeling the sequence of arguments in a discussion thread on ``Change My View (CMV)'', a discussion forum on Reddit. \cite{wang-etal-2019-persuasion} has investigated which types of persuasion strategies
have a more significant impact in convincing people to donate to a specific charity. This work is the first step in building personalized persuasive dialogue systems. Furthermore, to study whether particular types of people find
particular argument styles more convincing, \cite{lukin2017argument} has collected a new corpus of personality information and belief change in socio-political arguments. They have shown that belief change is affected by personality factors. For example, conscientious people are more convinced by dialogic, emotional arguments, while agreeable people are more likely to be persuaded by dialogic, factual arguments.
Inspired by this line of research, this dissertation further investigates the effect of source and audience factors in persuasion by asking the following questions:

\begin{enumerate}
    \item  How do the prior beliefs of the audience affect the process of persuasion?
    \item  Do social interactions play an essential role in people's success in online argumentation?
    \item How can we measure the relative impact of source and audience factors, language, and the pragmatic context in computational studies of persuasion and argument impact prediction? 
\end{enumerate}

Prior work has also investigated the related tasks of argument quality assessment and argument impact prediction \citep{persing-ng-2015-modeling, el-baff-etal-2018-challenge, wachsmuth-etal-2017-argumentation}. For example, \cite{persing-ng-2015-modeling} has introduced a corpus of argumentative student essays annotated with argument strength scores. They have further proposed a supervised, feature-based model to score the essays based on argument strength automatically. \cite{wachsmuth-etal-2017-computational} has studied logical, rhetorical, and
dialectical quality dimensions and proposed a taxonomy of argumentation quality from these dimensions. \cite{el-baff-etal-2018-challenge} has explored argument quality in news editorials, collecting annotations for the perceived effect of editorials from the New York Times. We have further explored the role of pragmatic context in predicting the perceived impact of arguments on online argumentation platforms \citep{durmus-etal-2019-role}.

\section{Argument Generation}

Argumentation is a significant part of a wide range of human activities. Humans are constantly confronted by situations where they are trying to persuade or are being persuaded. A major goal of computational argumentation is to build systems that can have meaningful debates and argumentative interactions with humans. Recent work in the area has made progress towards this goal through the automated generation of argumentative text \citep{zukerman-etal-2000-using, sato-etal-2015-end,hua-wang-2018-neural, bar-haim-etal-2020-arguments}. \cite{zukerman-etal-2000-using} and \cite{alshomary-etal-2020-target} have proposed a Bayesian argument generation system to generate arguments given the corresponding argumentation strategies. \cite{sato-etal-2015-end} has presented a sentence-retrieval-based end-to-end argument generation system that can participate in English debating games. \cite{hua-etal-2019-argument} has explored a neural counter-argument generation method that consists of a text planning decoder and a content realization decoder to select the main talking points and generate an argument given the talking points.  \cite{hidey-mckeown-2019-fixed} has further proposed a neural model that edits the original claim semantically to produce a claim with an opposing stance. Similarly, \cite{hua-wang-2018-neural} has studied the task of generating arguments of a different stance for a given argument. They have further incorporated external knowledge into the encoder-decoder architecture and have shown that their model can generate arguments that are more likely to be on topic. \cite{wang-ling-2016-neural} and \cite{bar-haim-etal-2020-arguments} have investigated the problem of summarizing the key points of an argument. Most recently, \cite{PMID:33731946_debater} has proposed an autonomous debating system (Project Debater) that can engage in a competitive debate with humans by generating a pipeline of four main modules: argument mining, an argument knowledge base (AKB), argument rebuttal, and debate construction. They have shown that their debating system can engage in a competitive debate with humans. However, they highlight the difficulty of achieving this end-goal due to the following reasons: 
\begin{enumerate}
    \item The outcome of the debates (i.e., selection of the winner) is highly subjective and open to interpretation since it may dependend on the characteristics of the audience. 
    \item Unlike other games such as chess \citep{CAMPBELL200257} or backgammon \citep{Tesauro1995}, humans would expect to be able to interpret every move of the system since they vote to decide the winner of the debate.
    \item There are a limited number of structured debate datasets to train such systems. 
\end{enumerate}

\chapter{A Dataset for Modeling User Characteristics in Persuasion}  \label{dataset}

Previous work in Natural Language Processing (NLP) and Computational Social Science (CSS) that studies persuasion has mainly focused on identifying the content and structure of an argument \citep{feng2011classifying} along with the linguistic features that are indicative of effective argumentation strategies \citep{10.1145/2872427.2883081}. However, the effectiveness of an argument cannot be determined solely by its textual content; instead, it is essential to consider the reader's or participant's characteristics in the debate or discussion. Does the reader already agree with the argument's stance?  Is she predisposed to changing her mind on a particular topic?  Is the style of the argument appropriate for the individual? To date, existing argumentation datasets have permitted only a limited assessment of such ``user" traits because information on the background of users is generally unavailable. This chapter introduces a new dataset with a wide range of user information to make progress towards this goal. We view this new dataset as a resource that allows the NLP and CSS communities to understand the effect of
audience characteristics on the efficacy of different persuasion strategies. In the following subsection, we describe our dataset in the context of existing argumentation datasets. We then provide a description and statistics for the key aspects of the dataset.

\section{Related Work and Datasets} \label{related_dataset}

There has been a tremendous amount of research effort to understand the important linguistic 
features for identifying argument structure and determining effective argumentation strategies
in a monologic
text \citep{article_moens,feng2011classifying,Stab2014IdentifyingAD,guerini-etal-2015-echoes}. For example, \citet{habernal-gurevych-2016-makes} has experimented with different machine learning models to predict which of the two given arguments is more convincing. To understand what kind of persuasive strategies are 
effective, \citet{DBLP:conf/argmining/HideyMHMM17} has further annotated different modes of persuasion (i.e., ethos, logos, 
pathos) and looked at which combinations appear most often in more persuasive arguments. 

Understanding argumentation strategies in conversations and the effect of the interplay between the participants' language has also been an important avenue of research. \cite{10.1145/2872427.2883081}, for example,  has examined the effectiveness of arguments
on ChangeMyView\footnote{\href{https://www.reddit.com/r/changemyview/} {https://www.reddit.com/r/changemyview/}.}, a debate forum in which people invite others
to challenge their opinions. They found that the interplay between the language of the opinion holder and that of the counterargument provides highly predictive cues of persuasiveness. \cite{zhang2016conversational} has examined the effect of conversational style in Oxford-style debates and found that the side that can best adapt in response to opponents' discussion points throughout the debate is more likely to be more persuasive.   

Although research on computational argumentation has mainly focused on identifying important
linguistic features 
of the argument, there is also evidence that it is important to model 
and account for the information about 
the debaters and the people who are judging the quality of the arguments: 
multiple studies in Social Sciences and Psychology show that people perceive arguments from different perspectives depending on their backgrounds and experiences \citep{correll2004affirmed,hullett2005impact,petty1981personal,lord1979biased,vallone1985hostile}. \citet{lukin2017argument} is one of the first to computationally study the impact of the audience, looking at the effect of their {\sc ocean} personality traits \citep{doi:10.1177/0146167202289008,norman_1963} on how they judge the persuasiveness of monologic arguments. This work is the most similar to ours since the effect of users' personalities is explored in the persuasion process. Our dataset does not have explicit information about users' personality traits; however, we have extensive information about their demographics, social interactions, beliefs, and language use. \citet{durmus-cardie-2019-corpus} describes the details of this dataset.

\section{DDO Dataset}

We collected $78,376$ debates 
from debate.org (DDO)\footnote{\href{http://www.debate.org}{www.debate.org}} from $23$ different topic
categories, including \textit{Politics}, \textit{Religion}, \textit{Health}, \textit{Science}, and \textit{Music}.\footnote{The dataset is publicly available at \href{http://www.cs.cornell.edu/~esindurmus/}{http://www.cs.cornell.edu/~esindurmus/}.} DDO  is an online argumentation platform where people can engage in debates, participate in forums and polls, and post their opinions on controversial topics. Participating in debates provides users an opportunity to challenge other users to change their opinions. After participating in debates, they receive feedback from the audience on the platform. This feedback mechanism is helpful for users to develop strategies to improve their debating skills over time. In addition to the text of the debates, we collected votes from the readers of these debates. Votes evaluate different dimensions of the debate, and they are important to determine which debaters are more successful in persuading other users. 

Each user creates a profile on this platform to share information about their background and preferences. To study the characteristics of users on persuasion, we collected user information for $45,348$ different users. In the next section, we share more details about the debates and the user information on this dataset. 

\subsection{Debates} \label{debates}

\begin{figure*}[t]
\centering
\includegraphics[scale=1]{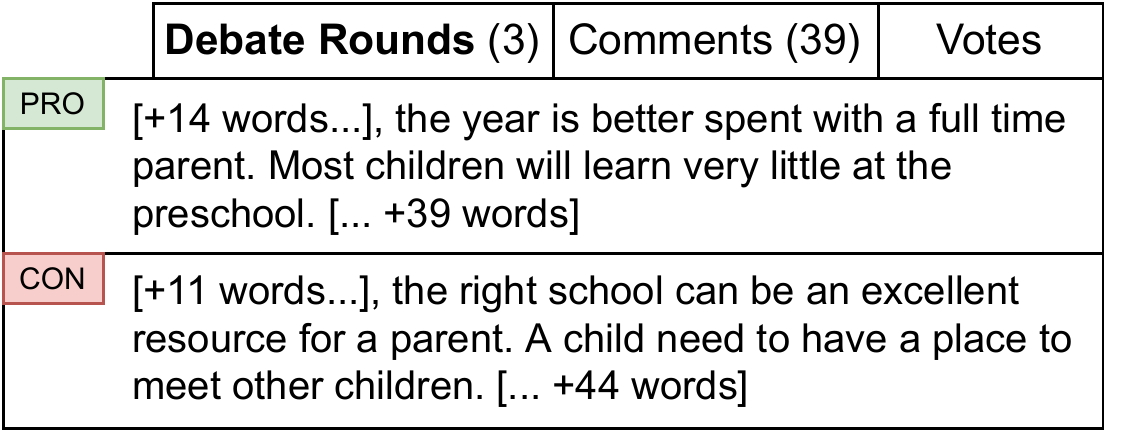}
\caption{{\sc round 1} for the debate claim \sc {``Preschool Is A Waste Of Time.".}}
\label{figure_round}
\end{figure*}

\hspace{1em}\textbf{Debate rounds.}
Each debate consists of a sequence of {\sc rounds} in which two debaters from opposing sides (one is supportive of the claim (i.e., {\sc pro}) and the other is against the claim (i.e., {\sc con})) provide their arguments. Each debater has a single chance in a {\sc round} to make their points. Figure \ref{figure_round} shows an example {\sc round 1}  for the debate claim {\sc ``Preschool Is A Waste Of Time"}. 
The number of {\sc rounds} in a debate ranges from $1$ to $5$, and most debates contain $3$ or more {\sc rounds}. The goal of the debaters in each {\sc round} is to provide arguments that would refute the opponent's points and convince readers to side with their stance.

\begin{figure*}[t]
\centering
\includegraphics[scale=0.8]{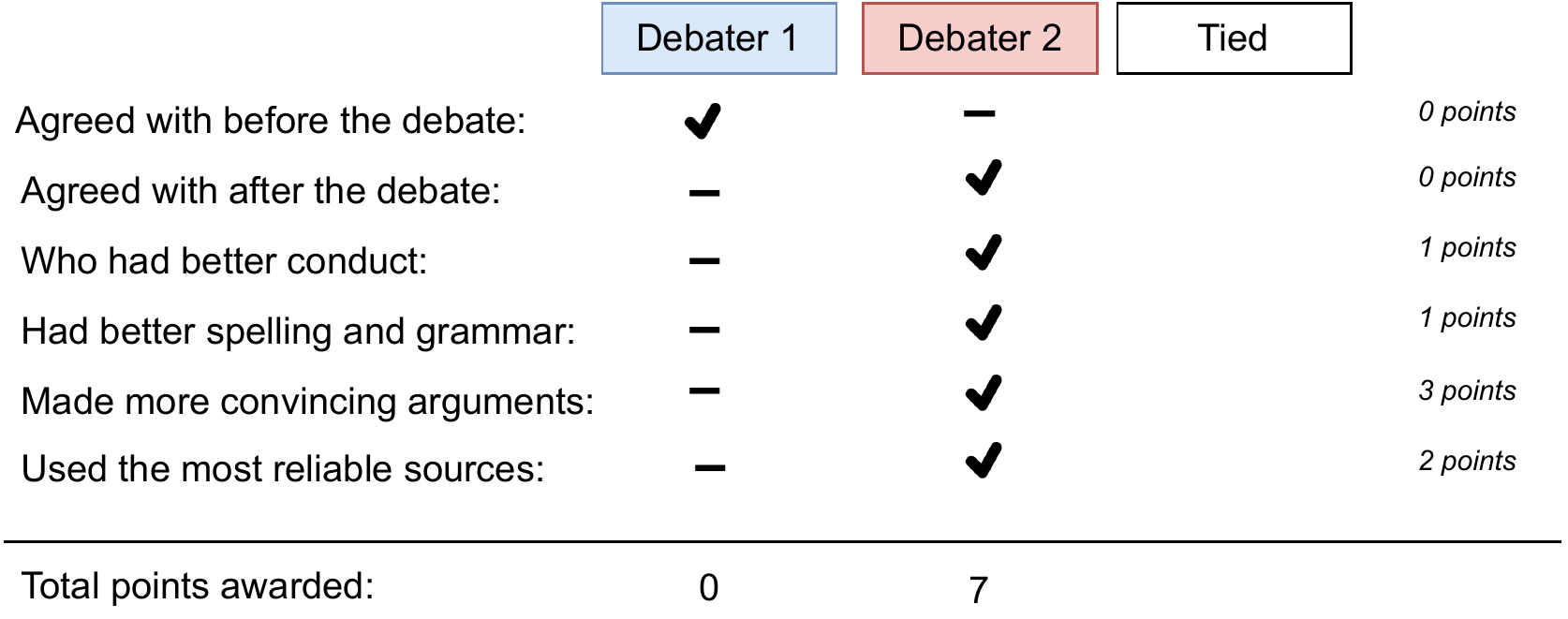}
\caption{An example post-debate vote.}
\label{figure_vote}
\end{figure*}

\textbf{Votes.} 
All users in the \textit{debate.org} community can vote on debates. As shown in Figure \ref{figure_vote}, voters share their stances on the debate topic before and after the debate and evaluate the debaters' conduct, spelling and grammar, persuasiveness, and reliability of the sources they refer to. For each such dimension, voters can choose one of the debaters as better or indicate a tie. 
The audience scores the debaters on these different aspects, and a winner is declared accordingly.
\footnote{Having better conduct: 1 point, having better spelling and grammar: 1 point, making more convincing arguments: 3 points, using the most reliable sources: 2 points.} This fine-grained voting system gives a glimpse into the reasoning behind the voters' decisions. \\

\begin{figure*}[t]
\centering
\includegraphics[scale=0.9]{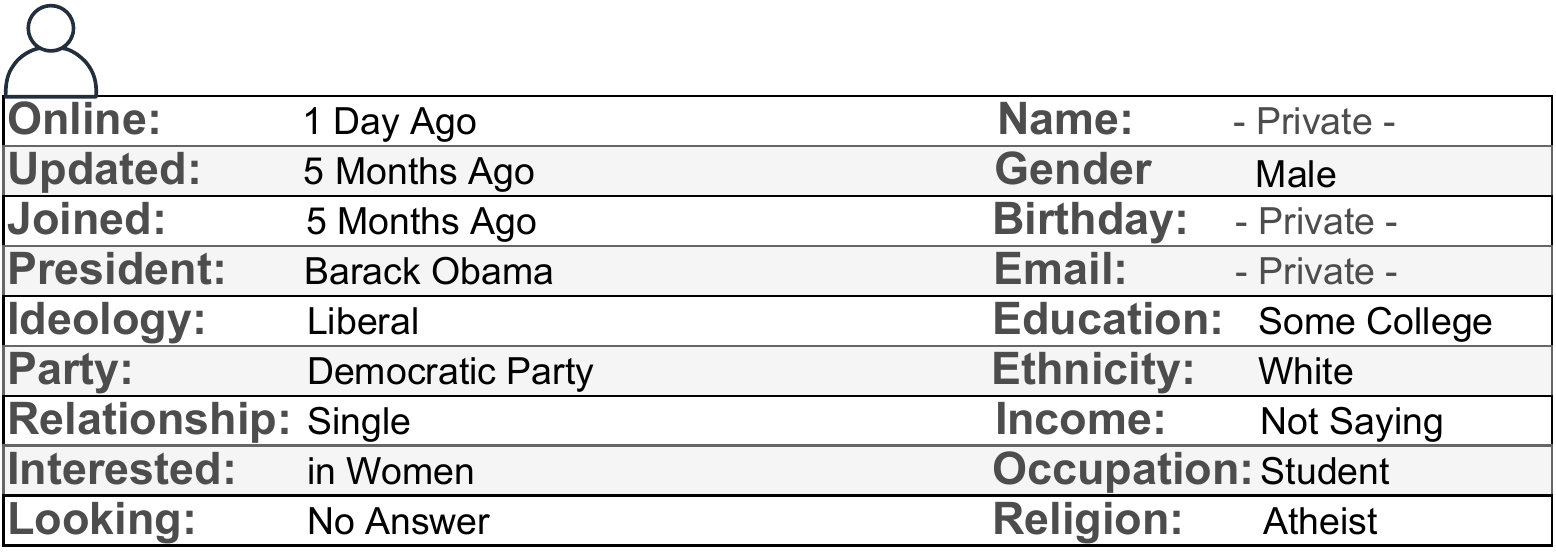}
\caption{Demographic and private state information for an example user profile.}
\label{figure_user_state}
\end{figure*}

\subsection{User information} 

The dataset includes extensive information about the users' demographics and private state, their activity on this platform, and their stance on various controversial topics. In this section, we describe the user information that is available in this dataset.

\subsubsection{Demographic and Private State Information}
On \textit{debate.org}, each user has the option to share demographic and private state information such as their age, gender, ethnicity, political ideology, religious ideology, income level, education level, and the political party they support. Figure \ref{figure_user_state} provides an example for the demographic and state information included in a user profile. We can see that these users select their political ideology, ethnicity, education, religious ideology, etc. However, they prefer not to share some of the information about themselves, such as their birthday, email, and income level, since sharing the demographic and state information is optional. 

\begin{figure*}[t]
\centering
\includegraphics[scale=0.85]{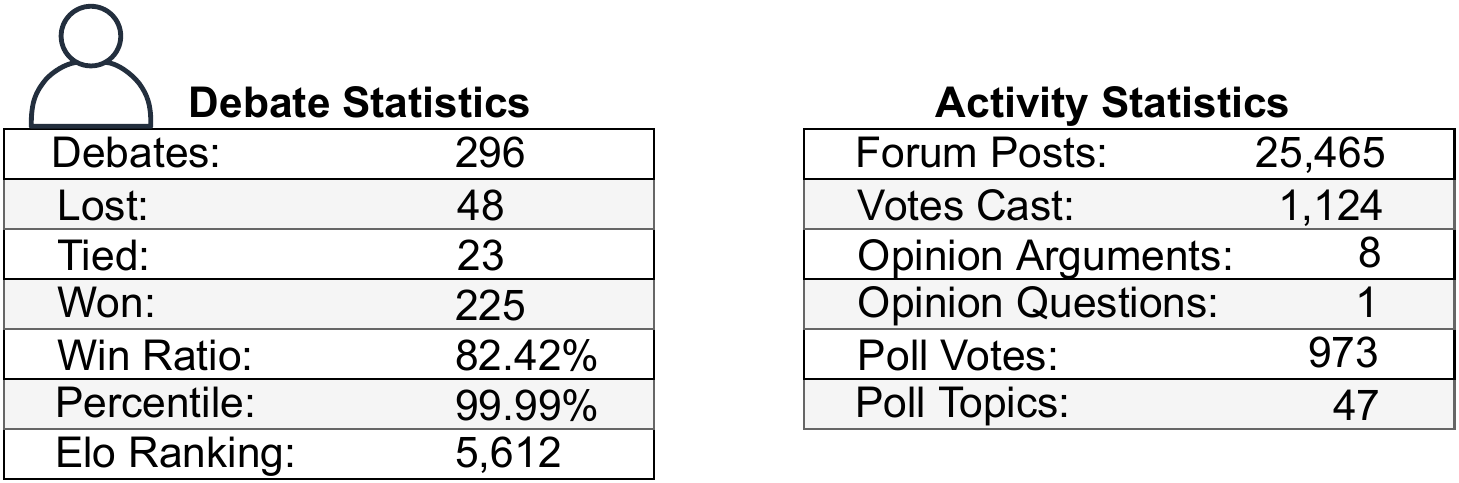}
\caption{Information about the activities of an example user profile.}
\label{figure_user_activity}
\end{figure*}

\subsubsection{User Activity Information}

Beyond the demographic and private state information, we have access to information about their activities on the website, such as their debating success rate, their participation both as debaters and voters, their votes, their forum posts, opinion arguments, opinion questions, poll votes, and poll topics that they created. The activities of an example user is shown in Figure \ref{figure_user_activity}. The availability of this information provides an opportunity to study users' interactions and success on this platform over time. 

\begin{figure*}[t]
\centering
\includegraphics[scale=0.85]{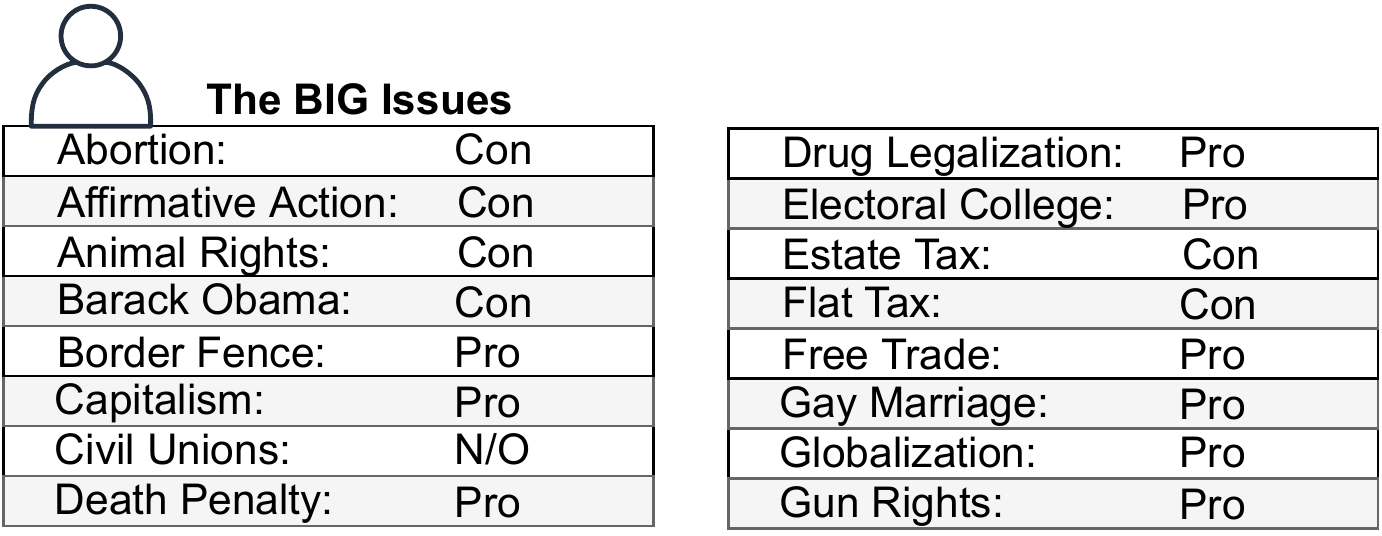}
\caption{Opinions on the \textbf{Big Issues} of an example user profile.}
\label{figure_user_big_issues}
\end{figure*}

\subsubsection{User Opinions on the \textit{Big Issues}} \label{big_issues_description}

The editors of the platform determine a list of the most controversial debate topics. These are referred to as \textit{big issues}\footnote{\label{big_issues}\href{http://www.debate.org/big-issues/}{http://www.debate.org/big-issues/}}. Each user has the option to share their stance on each \textit{big issue} on their profile (see Figure \ref{figure_user_big_issues}): either {\sc pro} (in favor), {\sc con} (against), {\sc n/o} (no opinion), {\sc n/s} (not saying), or {\sc und} (undecided). This gives a glimpse into the prior stance of users on a wide range of controversial topics. Moreover, this information can be used to determine opinion similarity between a pair of users. 

\section{Data Statistics}

\begin{figure}[]
\centering
\includegraphics[scale=0.65]{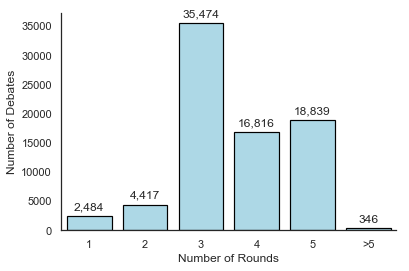}
\caption{The number of debates with the given number of rounds.}
\label{figure_num_rounds}
\end{figure}

\begin{figure}[]
\centering
\includegraphics[scale=0.65]{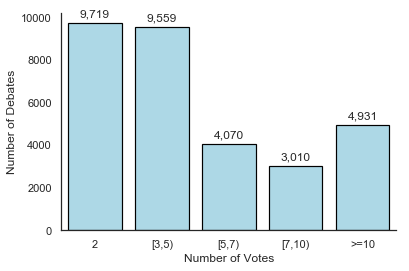}
\caption{The number of debates for a given range of votes.}
\label{figure_num_votes}
\end{figure}

The dataset consists of 78,376 debates from October of 2007 until November of 2017 with comprehensive user profile information for 45,348 users. Statistics on the number of debates with their corresponding number of rounds and votes are shown in Figure \ref{figure_num_rounds} and Figure \ref{figure_num_votes}, respectively. The majority of debates have 3 to 5 rounds. There are some debates with only one round; however, most debates have two or more rounds since the debates are highly interactive.

Although there are many debates with no votes, around 21k debates have three or more votes. We disregard the debates with $\leq3$ votes in our studies in order to have enough feedback to model the factors of success in persuasion. 

\begin{figure}[]
\centering
\includegraphics[scale=0.7]{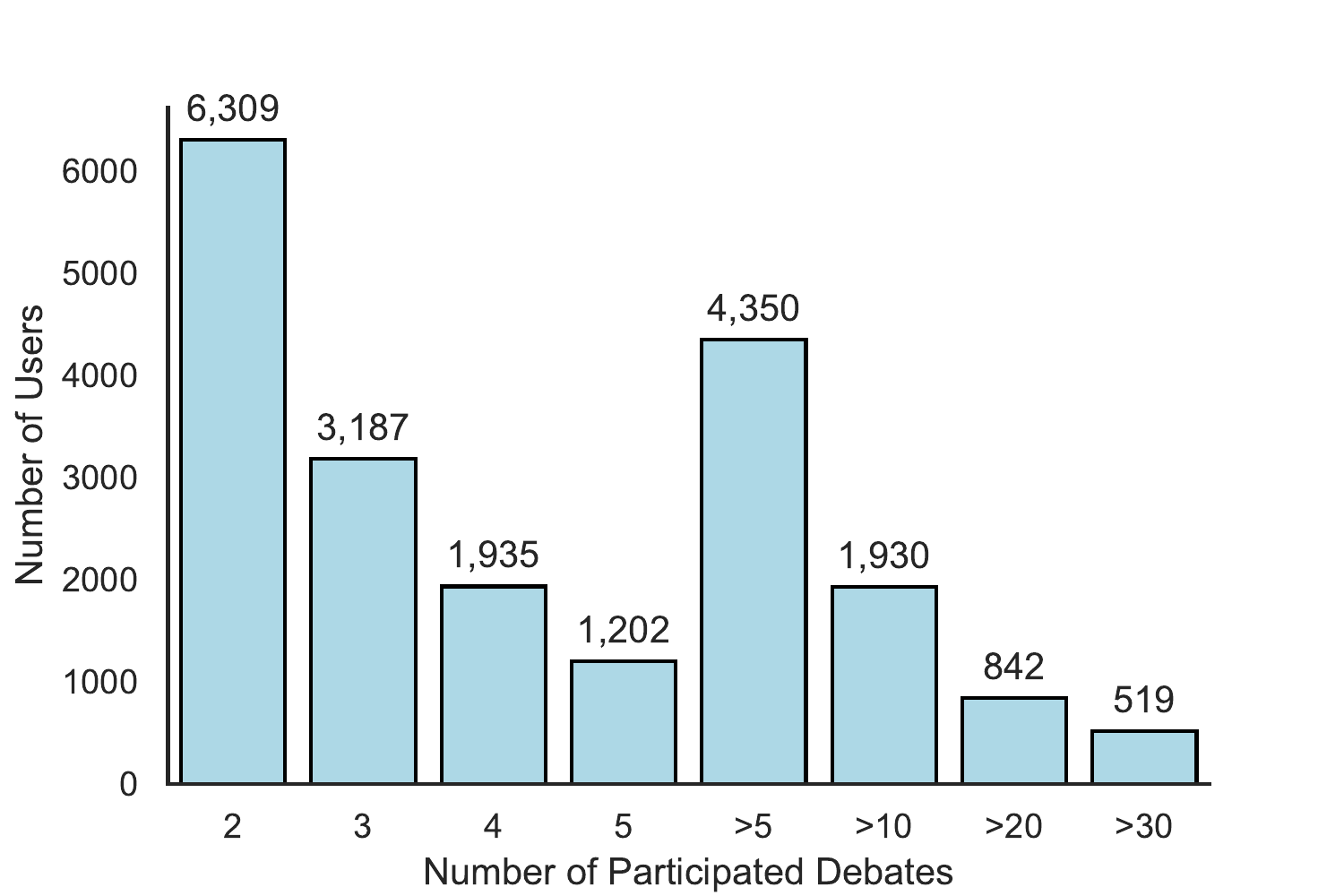}
\caption{The number of users that have participated in a given number of debates.}
\label{figure_user_num_deb}
\end{figure}

Figure \ref{figure_user_num_deb} shows the number of debates that users participated in. The majority have participated in only a single debate. However, some users actively participate in many debates. For example, around 2k debaters have participated in more than ten debates during the period included in the dataset. We study these debaters to understand the factors of debating success over time. 

\section{Limitations}
The dataset includes comprehensive information about users on the platform, which allows us to model user factors in persuasion. However, we acknowledge that we are unable to represent all demographics due to a lack of data. Participation on the platform tends to be highly skewed towards an American audience. Moreover, even within this group, the distribution of user characteristics may not be representative enough. Therefore, some valid opinions may be under-represented, and this should be accounted for while employing models derived from this data. Furthermore, we assume that the information users share on their profiles is accurate, and we use this information to model their characteristics. However, there is no mechanism on this platform to ensure that users provide accurate information. 

\section{Chapter Summary}

In this chapter, we present a novel dataset, DDO, of debates collected from \href{{debate.org}}{\textit{debate.org}}. The dataset includes interactive debates along with votes from the audience to evaluate various aspects of each debater. Moreover, the dataset has comprehensive information about the users on the platform. This allows us to study the effect of source and audience factors in persuasion (Chapter \ref{prior_beliefs}). We further use this dataset to model the impact of social interactions on long-term success in online debating (Chapter \ref{social_interactions}).

\chapter{The Role of the Speaker and the Audience Factors in Computational Persuasion} \label{prior_beliefs}

Using the DDO dataset described in Chapter \ref{dataset}, this chapter studies the effect of factors associated with the speaker and the audience of a debate to assess the more persuasive debater with respect to an individual audience member.

\section{Background}
Most of the recent work in computational persuasion has focused on identifying the characteristics of persuasive language \citep{habernal-gurevych-2016-makes, DBLP:conf/argmining/HideyMHMM17}. However, there is evidence from the Social Sciences and Psychology that non-content cues such as the factors of the speaker and the audience play an essential role in persuasion and opinion formation. Instead of carefully processing the content of the arguments, people may rely on simple non-content heuristics in decision making \citep{edsovi.00005205.197911000.0001619791101}. Understanding the effect of persuasion strategies on people, the biases people have, and the impact of people's prior beliefs on their opinion change has been an active area of research \citep{correll2004affirmed,hullett2005impact,petty1981personal}. 

Prior work has shown that the speaker's credibility is an essential factor for people's perceptions of the arguments \citep{edsgcl.1760738119951001, edsovi.00005205.198011000.0000119801101}. For example, there is a significant correlation between the communication speed and the persuasive effect of the arguments. The audience perceived a communicator with a faster communication rate as more credible without really focusing on the content of the arguments \citep{mcguire1985attitudes,edsgcl.1760738119951001}. Furthermore, \citet{edsovi.00005205.198011000.0000119801101} has studied the effect of a communicator's perceived likability in opinion formation and found that low-involvement subjects perceive the arguments of likable communicators as more persuasive. High involvement subjects (i.e., the subjects who feel their opinion judgments have essential consequences
for themselves) have shown to have a more systematic strategy that assigns a higher weight to the message content in opinion formation. A communicator's perceived attractiveness is also positively correlated with their persuasiveness since the audience perceives more attractive communicators as more effective \citep{ 1980-32482-00119790801, eagly1975attribution}.  

There is further evidence showing that people's prior beliefs significantly affect their opinion formation \citep{edsbds.17414237419960101}. People with strong prior beliefs on controversial issues have shown to have biased stances even when they are presented with empirical evidence: i.e., they tend to find empirical evidence that is confirming their prior beliefs more convincing \citep{edsovi.00005205.197911000.0001619791101}.  
Similarly, people judge the fairness and reliability of source content in a biased way; i.e., they accept evidence that supports their stance at face value while scrutinizing evidence that threatens their initial position \citep{vallone1985hostile}. Inspired by these findings, we study the impact of prior beliefs in computational persuasion in this chapter. 
\citet{lukin2017argument} is the most relevant work to ours since they investigated the effect of an individual's personality features (open, agreeable, extrovert, neurotic, etc.) on the type of argument (factual vs.\ emotional) they find more persuasive. Our work differs from this work since we study debates. In addition, we look at different types of user profile information, such as a user's religious and ideological beliefs and prior beliefs and opinions of the audience on various topics \citep{durmus-etal-2019-role, longpre-etal-2019-persuasion}. 

\section{Role of Prior Beliefs in Computational Persuasion}

Using the DDO dataset (described in Chapter \ref{dataset}), we first analyze which dimensions of argument quality are the most important for determining the successful debater. Then, we investigate whether there is any connection between selected user-level factors and users' opinions on the \textit{big issues} to see if we can infer their opinions from these factors. Finally, using our findings from these analyses, we perform the task of predicting which debater will be perceived as more successful by a given voter. In this study, we particularly aim to understand the role of users' prior beliefs (i.e., their self-identified political and religious ideology) in predicting the more successful debater.

\subsection{Relationships between argument quality dimensions}

In Section \ref{debates}, we describe the aspects the voters evaluate in order to determine which debater is more successful.
There are two alternative criteria for determining the successful debater. We consider both in our experiments.

\textbf{Criterion 1: Argument quality.}
Debaters get points for each dimension of the debate. The most important dimension --- in that it contributes most to the point total --- is making convincing arguments. The debater with the highest point total is declared the winner. \textit{debate.org} uses Criterion 1 to determine the winner of a debate. 

\textbf{Criterion 2: Convinced voters.} Alternatively, since voters share their stances before and after the debate, the debater who convinces more voters to change their stance can be considered the winner.

\begin{figure*}[h]
\centering
\includegraphics[scale=0.65]{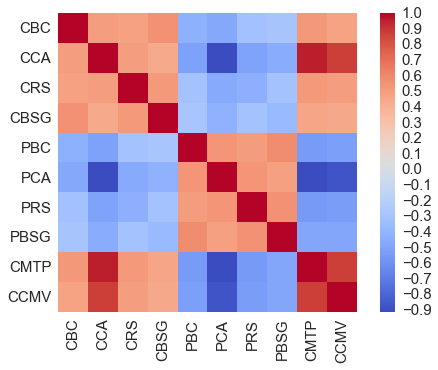}
\caption{The correlations among argument quality dimensions.}
\label{voting_dimension}
\end{figure*}

Figure \ref{voting_dimension} shows the correlation between pairs of voting dimensions (in the
first eight rows/columns) along with the correlation of each dimension with (1) getting the highest point total 
(row/column 9) and (2) convincing more to change their stance (final row/column).
The abbreviations in Figure \ref{voting_dimension} stand for (on the {\sc con} side): has better conduct ({\sc cbc}), 
makes more convincing arguments ({\sc cca}), uses more 
reliable sources ({\sc crs}), has better spelling and grammar ({\sc cbsg}), gets more total points ({\sc cmtp}) and
convinces more voters ({\sc ccmv}). For the {\sc pro} side we use {\sc pbc}, {\sc pca}, and so on. 

From Figure \ref{voting_dimension}, we can see that making more convincing arguments ({\sc cca}) correlates the most with total points ({\sc cmtp}) and
convincing more voters ({\sc ccmv}). This suggests that the language of the argument is important in persuading the audience, and it motivates us to identify the linguistic features that are indicative of convincing arguments while taking into account speaker and audience factors.

\subsection{The relationship between a user's opinions on the \textit{big issues} and their prior beliefs}
\begin{figure*}
\centering
\includegraphics[scale=0.7]{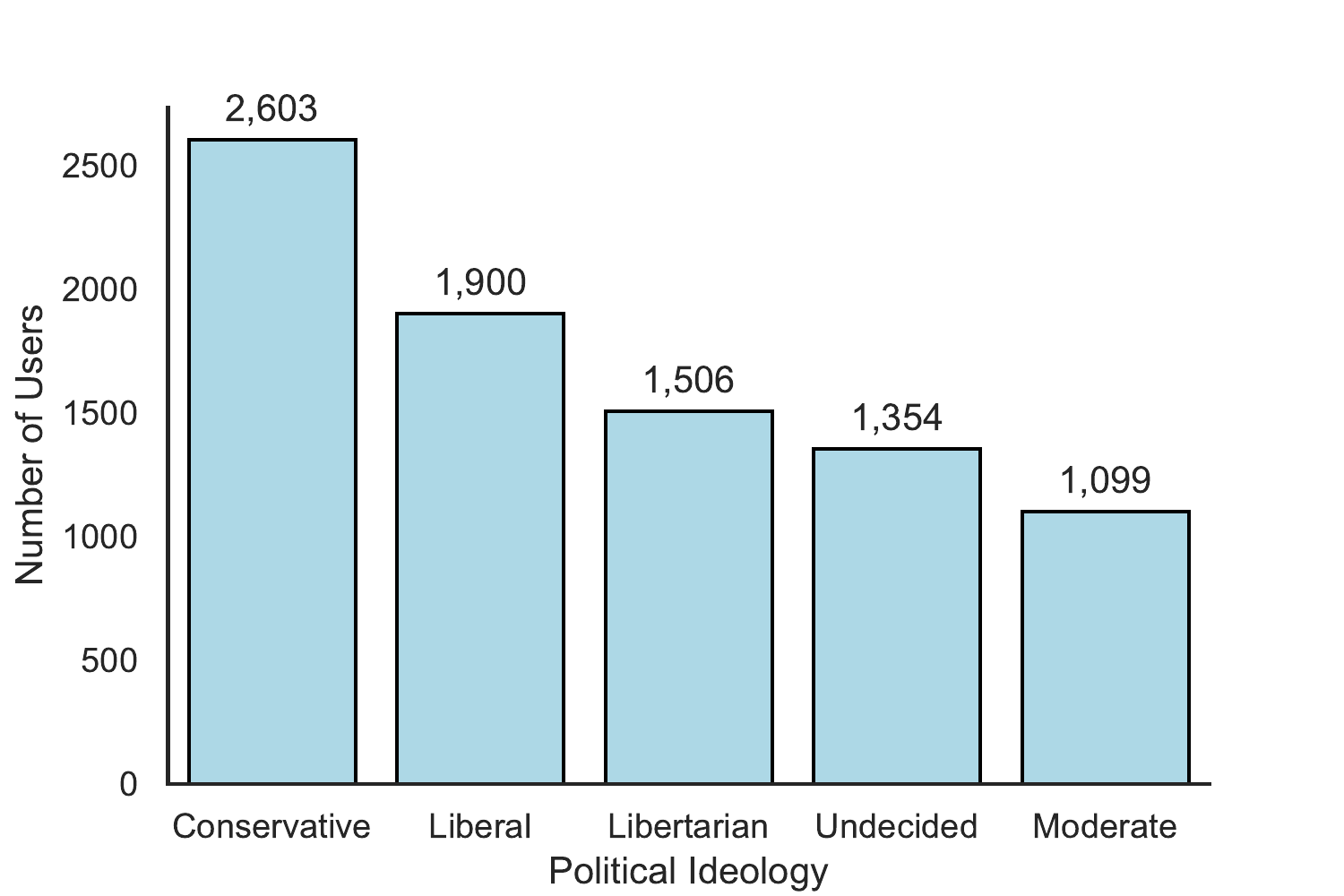}
\caption{The number of users with the given political ideology.}
\label{user_stats_pol_ideology}
\end{figure*}

\begin{figure*}[h]
\centering
\includegraphics[scale=0.7]{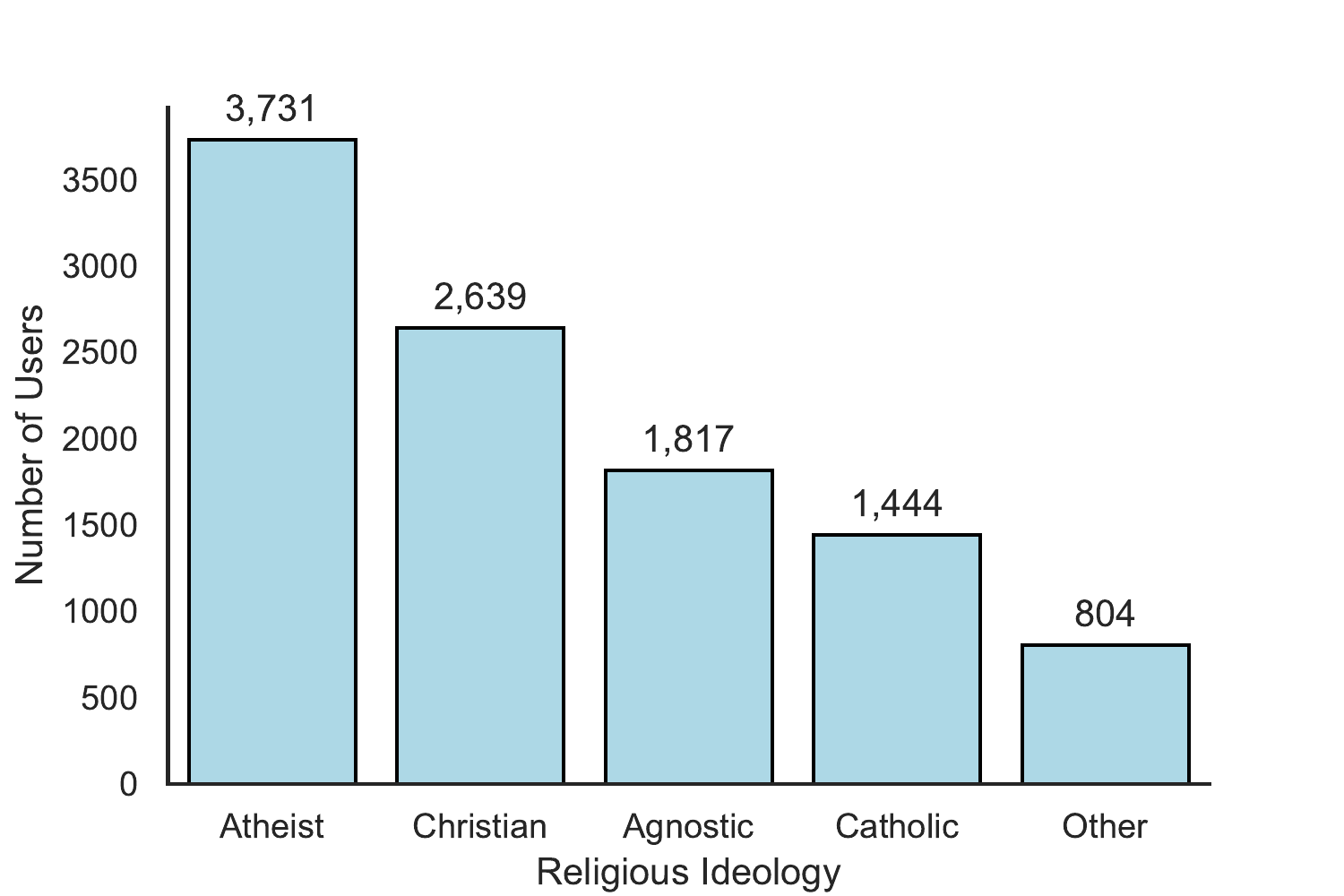}
\caption{The number of users with the given religious ideology.}
\label{user_stats_rel_ideology}
\end{figure*}

As described in Section \ref{big_issues_description}, users share their self-identified political and religious ideologies along with their opinions on various controversial issues (i.e., \textit{big issues}). Note that many people prefer not to share their political and religious ideologies. 
Figures \ref{user_stats_pol_ideology} and \ref{user_stats_rel_ideology} show the number of users who self-identify with the given political or religious ideology.

We disentangle different aspects of a person's prior beliefs in order to understand how they correlate with their opinions on the \textit{big issues}. We focus on prior beliefs in the form of self-identified political and religious ideology.

\begin{figure*}
\centering
\includegraphics[scale=0.49]{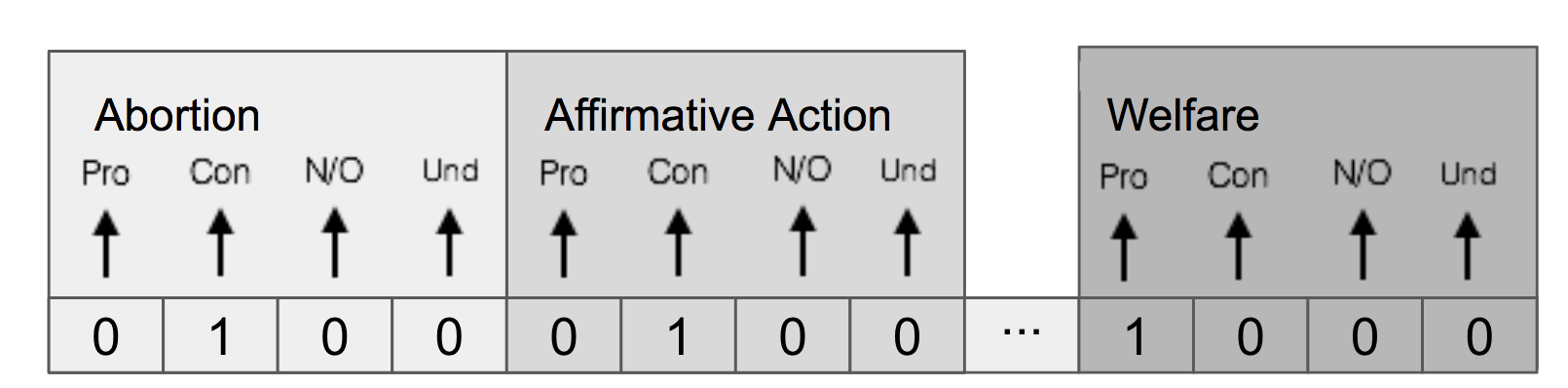}
\caption{The representation of the {\sc BigIssues} vector derived by this user's decisions on Big Issues. Here,  the user is {\sc con} for {\sc abortion} and {\sc affirmative action} issues and {\sc pro} for the {\sc welfare} issue.}
\label{figure_big_issues}
\end{figure*}

\textbf{Representing the \textit{big issues}.} To represent a user's opinion on a particular \textit{big issue}, we use a four-dimensional one-hot encoding where the indices of the vector correspond to {\sc pro}, {\sc con}, {\sc n/o} (no opinion), and {\sc und} (undecided), consecutively. Note that we do not have a representation for {\sc n/s} (not saying) since we eliminate users who indicate {\sc n/s} for any of the  \textit{big issues}.  We then concatenate the vector for each of the \textit{big issue} to represent a user's stance on all the \textit{big issues} as shown in Figure \ref{figure_big_issues}. We denote this vector by {\sc BigIssues}.

We test the correlation between an individual's opinion on the \textit{Big Issues} and the selected user-level factors in this study using two approaches: clustering and classification.

\begin{figure*}[h]
  \centering
  \subfigure[{\sc liberal} vs. {\sc conservative}
  ]{\includegraphics[scale=0.7]{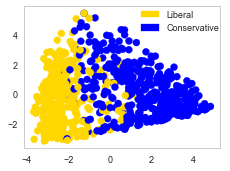}.}
  \quad
  \subfigure[{\sc atheist} vs. {\sc christian}.]
  {\includegraphics[scale=0.7]{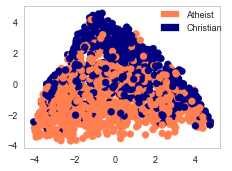}}
  \caption{PCA representation of decisions on Big Isues color-coded with political and religious ideology. We see more distinctive clusters for {\sc conservative} vs. {\sc liberal} users suggesting that people's opinions are more correlated with their political ideology.}
\label{clustering_figure}
\end{figure*}

\textbf{Clustering the users' decisions on the \textit{big issues}.} We apply PCA on the {\sc BigIssues} vector of users who identified themselves as {\sc conservative} vs.\ {\sc liberal} ($740$ users). We do the same for the users who identified
 themselves as {\sc atheist} vs.\ {\sc christian} ($1501$ users).
 In Figure \ref{clustering_figure}, we see distinct clusters of {\sc conservative} vs.\ {\sc liberal} users in the two-dimensional representation, while for {\sc atheist} vs.\ {\sc christian}, the separation is not as distinct. This suggests that people's opinions on the \textit{big issues} identified by \textit{debate.org} correlate more with their political ideology than their religious ideology.

\textbf{Classification approach.} We can also treat this as a classification task\footnote{For all the classification tasks described in this paper, we experiment with logistic regression, optimizing the regularizer ($\ell$1 or $\ell$2) and the
regularization parameter C (between $10^{-5}$ and $10^{5}$).} using the {\sc BigIssues} vector for each user as the input feature and the user's religious and political ideology as the labels to be predicted.
Table \ref{belief_table}
shows the prediction accuracy for religious and political ideology. We see that using the {\sc BigIssues} vector as a feature performs
significantly better\footnote{\label{significance}We performed the McNemar significance test.} than the majority 
baseline.\footnote{The majority class baseline predicts {\sc conservative} for political and {\sc christian} for religious ideology for each example, respectively.}

\begin{table}
\begin{center}
\begin{tabular}{|l|rl|}
\hline \bf Prior belief type & \bf Majority & \bf \sc{BigIssues} \\ \hline
Political Ideology & $57.70$\% & $92.43$\% \\
Religious Ideology & $52.70$\% & $82.81$\% \\
\hline
\end{tabular}
\end{center}
\caption{\label{font-table} Accuracy using majority baseline vs.\ {\sc BigIssues} vectors as features.}
\label{belief_table}
\end{table}

This analysis shows a clear relationship between people's opinions on the \textit{big issues} and the selected user-level factors. 
It raises the question of whether it is even possible to persuade someone to change their stance on a given issue. It may be the case that people prefer to agree with the individuals with the same (or similar) beliefs regardless of the quality of opposing arguments.
Therefore, it is crucial to understand the relative effect of prior beliefs vs.\ argument strength on persuasion. 

\subsection{Task formulation}
Some of the previous work in NLP on persuasion, focuses on predicting the winner of a debate as determined by the change in the number of people supporting each stance before and after the debate \citep{zhang2016conversational,potash2017towards}. However, we believe that studies of the effect of language on persuasion should consider extra-linguistic factors that can affect opinion change. In particular, we propose an experimental framework for studying the effect of language on persuasion by controlling for the prior beliefs of the audience.
In order to do this, we formulate a more fine-grained prediction task: 
for a given voter, predict which 
side/debater/argument the voter will declare as the winner.

\textbf{Task 1: Controlling for religious ideology.}
In the first task, we control for religious ideology by selecting debates where the debaters have differing religious ideologies (e.g., debater 1 is {\sc atheist}, debater 2 is 
{\sc christian}). Also, we only consider voters that (a) self-identify with one of these 
religious ideologies (e.g., the voter is either {\sc atheist} or {\sc christian}) and (b) 
changed their stance on the topic after the debate.  
For each such voter, we want to predict which debater did the convincing. Thus, in this task, we use 
\textit{Criterion 2} to determine the winner of the debate from the voter's point of
view.  We hypothesize that a voter will be convinced by the
debater that espouses the religious ideology of the voter. 
Given this setting, we can study the factors that govern whether a debater can convince any given voter. It also provides an opportunity to understand how voters who change their minds perceive arguments from a debater with the same vs.\ opposing prior beliefs. 

To study the effect of the debate topic, we perform this study for two cases --- debates belonging to the \textit{Religion} category only vs.\ all categories. The \textit{Religion} category contains debates like {\sc ``Is the Bible against women's rights?'' } and {\sc ``Religious theories should not be taught in school''}. 
We expect to see a stronger effect due to prior beliefs for debates on \textit{Religion}.

\textbf{Task 2: Controlling for political ideology.}
Similar to the setting described above, Task 2 controls for political ideology. 
In particular, we only select debates where the debaters have differing political ideologies 
({\sc conservative} vs.\ {\sc liberal}). In contrast to Task 1, we consider all voters that 
self-identify with any of the debater's ideologies (regardless of whether the voter's
stance changed post-debate).  
%
For this task, we predict which debater will get assigned more points from a given voter.
Thus, Task 2 uses \textit{Criterion 1} to determine the winner of the debate 
from the point of view of a voter. We hypothesize that a voter will assign
more points to a debater who shares the same political ideology.

Similar to task 1, we perform the study for two cases --- debates from the \textit{Politics} 
category only and debates from all categories. We expect to see a stronger effect due to prior beliefs for debates on \textit{Politics}.

\begin{table*}[]
\centering
\begin{center}
\begin{tabular}{|p{6cm}p{8cm}|}
\hline
\textbf{User-based features}&\textbf{Description} \\
    \hline
    \textbf{Opinion similarity.} & For $userA$ and $userB$, the cosine similarity of {\sc BigIssues$_{userA}$} and {\sc BigIssues$_{userB}$}.\\
    \textbf{Matching features.} & For $userA$ and $userB$, 1 if $userA_f$==$userB_f$, 0 otherwise where
$f$ $\in$ $\{$political ideology, religious ideology$\}$. We denote these features as \textit{matching political ideology} and \textit{matching religious ideology}.\\
    \hline
    \textbf{Linguistic features} & \textbf{Description} \\
    \hline
    \textbf{Length.} &  Number of tokens.\\
    \textbf{Tf-idf.} & Unigram, bigram and trigram features. \\
    \textbf{Referring to the opponent.} &  Whether the debater refers to their opponent using words or phrases like ``opponent, my opponent''.\\
   \textbf{Politeness cues.} &  Whether the text includes any signs of politeness such as ``thank'' and ``welcome''.\\
    \textbf{Showing evidence.} & Whether the text has any signs of citing any other sources (e.g., phrases like ``according to''), or quotation.\\
    \textbf{Sentiment.} &  Average sentiment polarity. \\
    \specialcell[b]{\textbf{Subjectivity.}} &  Number of words with negative strong, ne-\\ 
    \citep{wilson2005recognizing} & negative weak, positive strong, and positive weak subjectivity. \\ 
    \textbf{Swear words.} & \# of swear words. \\
    \textbf{Connotation score } & Average \# of words with positive, negative  \\ 
    \citep{feng2011classifying} &   and neutral connotation.\\

    \textbf{Personal pronouns.} & Usage of first, second, and third person pronouns. \\
    \textbf{Modal verbs.} &  Usage of modal verbs.\\
    \textbf{Argument lexicon features.} & \# of phrases corresponding to different ar- \\
     \citep{somasundaran2007detecting} & gumentation styles. \\ 
    \textbf{Spelling.} &  \# of spelling errors.\\
    \textbf{Links.} & \# of links. \\
    \textbf{Numbers.} &  \# of numbers. \\
    \textbf{Exclamation marks.} &  \# of exclamation marks. \\
    \textbf{Questions.} &  \# of questions. \\
   
  \hline
\end{tabular}
\end{center}
\caption{Feature descriptions.}
\label{table:features}
\end{table*}

\subsection{Features} \label{features}
The features we use in our model are shown in Table~\ref{table:features}. They can be
divided into two groups --- features that describe the prior beliefs of the users
and linguistic features of the arguments.

\subsubsection*{User features}
We use cosine similarity between a voter and a debater's \textit{big issue} vectors. 
This feature gives an approximation of the overall similarity of two users' opinions.
We also use indicator features to encode whether the religious and political beliefs of a voter match that of a debater. 

\subsubsection*{Linguistic features}
We extract linguistic features separately for both the {\sc pro} and {\sc con} side of the debate (combining all the utterances of each side across the different turns). Table \ref{table:features} contains a list of these features. It includes features that carry information 
about the style of the language (e.g., usage of modal verbs, length, punctuation), represent different
semantic aspects of the argument (e.g., showing evidence, connotation \citep{feng2011classifying},
subjectivity \citep{wilson2005recognizing}, sentiment, swear word features) as well as
features that convey different argumentation styles (argument lexicon features \citep{somasundaran2010recognizing}. Argument lexicon features 
include the counts for the phrases that match various argumentation styles such as assessment, authority, conditioning, contrasting, emphasizing, generalizing, empathy, inconsistency, necessity, possibility, priority, rhetorical questions, desire, and difficulty. 
We then concatenate these features to get a single feature representation for the entire debate. 

\section{Results and Analysis}
\label{results}
For each of the tasks, prediction accuracy is evaluated using 5-fold cross-validation. We pick the model parameters for each split with 3-fold cross-validation on the training set. We do ablation for each of user-based and linguistic features. We report the results for the feature sets that perform better than the baseline. 


We perform analysis by training logistic regression models using only user-based features, only linguistic features, and finally combining user-based and linguistic features for both the tasks. 

\begin{table*}[]
\centering
\begin{tabular}{|p{7cm}|p{2cm}|}
\hline
		  & \textbf{Accuracy}  \\
\hline
\textbf{Baseline} &   \\

{Majority} & $56.10$\% \\ 
 \hline
\textbf{User-based Features} &   \\
      {Matching religious ideology} & \textbf{$65.37$} \% \\ 
    \hline
    \textbf{Linguistic features} &  \\
    {Personal pronouns} & \textbf{$57.00$} \%\\
    {Connotation} & \textbf{$61.26$} \%\\
    {All two features above} & \textbf{$65.37$} \%\\
    
   \hline
   	\textbf{User-based + Linguistic features} &   \\
   {{\sc user*} + Personal pronouns}  &$65.37$\%\\
   {{\sc user*} + Connotation}  &$66.42$\%\\
  {{\sc user*} + {\sc language*}}  &$64.37$\%\\
  \hline
\end{tabular}

\caption{Results for Task 1 for debates in category \textit{Religion}. 
{\sc user*} represents the best performing combination of user-based features. {\sc language*} represents the best performing combination of linguistic features.
Since using linguistic features only would give the same prediction for all voters in a debate, the maximum accuracy that can be achieved using language features only is $92.86$\%. }
\label{table:task1_religion}
\end{table*}

\textbf{Task 1 for debates in category \textit{Religion}.} As shown in Table \ref{table:task1_religion}, the majority baseline (predicting the winning side of the majority of training examples out of {\sc pro} or {\sc con}) gets $56.10$\% accuracy. User features alone perform significantly better than the majority baseline. The most important user-based feature is \textit{matching religious ideology}. This means it is very likely that people change their views in favor of a debater with the same religious ideology. In a linguistic-only feature analysis, the combination of the \textit{personal pronouns} and \textit{connotation} features emerges as most important and performs significantly better than the majority baseline with $65.37$\% accuracy. When we use both user-based and linguistic features, the accuracy improves to $66.42$\% with \textit{connotation} features. An interesting observation is that including the user-based features and the linguistic features changes the set of important linguistic features for persuasion, removing \textit{personal pronouns} from the important linguistic features set. This shows the importance of studying potentially confounding user-level factors.

\begin{table}[h]
\centering
\begin{center}
\begin{tabular}{|p{10cm}|p{2cm}|}
\hline
  & \textbf{Accuracy}  \\
\hline
\textbf{Baseline} &  \\
{Majority} & $57.31$\% \\ 
 \hline
\textbf{User-based Features} &   \\
{Matching religious ideology} & $62.79$ \% \\ 
 \specialcell[b]{Matching religious ideology + Opinion similarity} & \textbf{$62.97$}\%\\
      
    \hline
    \textbf{Linguistic features} &  \\
    {Length\footnote{This linguistic feature is the one achieving the best performance.}} &  $61.01$ \% \\
    \hline
   	\textbf{User-based + Linguistic features} &   \\
  {{\sc user*} + Length}  & $64.56$ \%\\
   \specialcell[b]{{\sc user*} + Length + Exclamation marks} & {$65.74$}\%\\
 \hline
\end{tabular}
\end{center}
\caption{Results for Task 1 for debates in all categories. 
The maximum accuracy that can be achieved using language features only is $95.77$\%.
}
\label{table:task_1_all}
\end{table}

\textbf{Task 1 for debates in all categories.} As shown in Table 
\ref{table:task_1_all}, for experiments with user-based features only, \textit{matching religious ideology} and \textit{opinion similarity} features are the most important. For this task, \textit{length} is the most predictive linguistic feature and can significantly improve the baseline ($61.01$\%). When we combine the language features with user-based features, we see that with \textit{exclamation mark}, the accuracy improves to ($65.74$\%).

\begin{table}[]
\begin{center}
\begin{tabular}{|p{8cm}|p{2cm}|}
\hline
& \textbf{Accuracy}  \\
\hline
\textbf{Baseline} &  \\ 
{Majority} & $50.91$\% \\ 
 \hline
\textbf{User-based Features} &   \\
  {Opinion similarity} & $80.00$ \% \\
    {Matching political ideology} & \textbf{$80.40$} \% \\
    \hline
    \textbf{Linguistic features} &  \\
    {Length} & $57.37$ \% \\
    \textit{linguistic feature set} & \textbf{$59.60$} \%\\
   \hline
   \textbf	{User-based + Linguistic features} &   \\
  {{\sc user*}+ \textit{linguistic feature set}}  & \textbf{$81.81$}\%
   \\
   \hline
\end{tabular}
\end{center}
\caption{Results for Task 2 for debates in category \textit{Politics}. 
The maximum accuracy that can be achieved using linguistic features only is $75.35$\%. 
The \textit{linguistic feature set} includes \textit{rhetorical questions, emphasizing, approval, exclamation mark, questions, politeness, referring to opponent, showing evidence, modals, links,} and \textit{numbers} as features.}
\label{table:task_2_politics}
\end{table}

\begin{table}[]
\begin{center}
\begin{tabular}{|p{8cm}|p{2cm}|}
\hline
 & \textbf{Accuracy}  \\
\hline
\textbf{Baseline} &   \\
{Majority} & $51.75$\% \\ 
 \hline
\textbf{User-based Features} &  \\
  {Opinion similarity} & 
 $73.96$\% \\
  
\hline
   
\textbf{Linguistic features} &  \\
    
{Length} & $56.88$\% \\
{Politeness} & $55.00$\% \\
{Modal verbs} & $52.32$\% \\
{Tf-idf features}  &  \textbf{$52.89$} \%\\
 \hline
 \textbf{User-based + Linguistic features} &   \\
 {{\sc user*}+ Length}  & $74.53$\%\\
 {{\sc user*}+ Tf-idf}  & $74.13$\%\\
 \specialcell[b]{{\sc user*}+ Length + Tf-idf} & \textbf{$75.20$}\%\\
  
\hline
\end{tabular}
\end{center}
\caption{Results for Task 2 for debates in all categories.
The maximum accuracy that can be achieved using linguistic features only is $74.53$\%.
}
\label{table:task_2_all}
\end{table}

\textbf{Task 2 for debates in category \textit{Politics}.} As shown in Table \ref{table:task_2_politics}, using user-based features only, the \textit{matching political ideology} feature performs the best ($80.40$\%). Linguistic features (refer to Table \ref{table:task_2_politics} for the full list) alone can still obtain significantly better accuracy than the baseline ($59.60$\%). The most important linguistic features include \textit{approval}, \textit{politeness}, \textit{modal verbs}, \textit{punctuation}, and \textit{argument lexicon features} such as \textit{rhetorical questions} and \textit{emphasizing}. 
When combining this linguistic feature set with the \textit{matching political ideology} feature, we see that accuracy improves ($81.81$\%). The \textit{length} feature does not improve when it is combined with the user features.

\textbf{Task 2 for debates in all categories.}
As shown in Table \ref{table:task_2_all}, when we include all categories, we see that the best performing user-based feature is the \textit{opinion similarity} feature ($73.96$\%). When using language features only, the \textit{length} feature ($56.88$\%) is the most important. For this setting, the best accuracy is achieved when we combine user features with \textit{length} and \textit{Tf-idf} features. We see that 
the set of language features that improves the performance of user-based features does not include some of the features that performed significantly better than the baseline when used alone (\textit{modal verbs} and \textit{politeness} features).

\section{Persuasion of the Undecided}

Research in psychology and political science suggests that there are critical differences in the persuasion of undecided versus decided voters/audience members.  For example, \citet{petty96} has found that prior experiences and beliefs can lead to the re-framing of a message perceived by a person to maintain consistency between their prior beliefs and their attitudes towards the topic of the message.
In particular, studies show that \textit{a priori} decided voters simply ignore certain information to maintain this consistency \citep{sween,vecc,kos14}. In contrast, an undecided voter is asked to decide on an issue for which previously received information was somehow unconvincing; and prior work has shown that, as a result, these voters are likely to rely heavily on information conveyed in a new message \citep{kos10, kos14, schill}.

Furthermore, the undecided voter group holds the highest potential for persuasion \citep{kos10,sheh17}. Public support for social and political causes often critically depends on the undecided decision-makers. Therefore, in our work, we explicitly study the factors that govern persuasion for \textit{a priori} \textsc{undecided} versus \textsc{decided} members of the audience \citep{longpre-etal-2019-persuasion}. 

\subsection{Task Formulation}

\begin{figure*}[h]
\centering
\includegraphics[width=14cm]{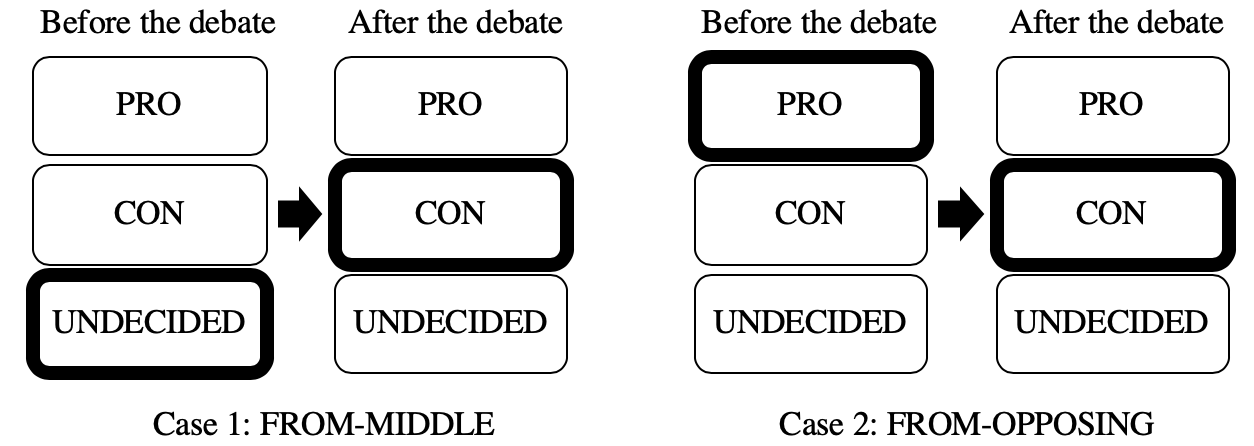}
\caption{\label{votes-ex} Example votes for a debate showing each case of persuasion. }
\end{figure*}

We aim to study the most important factors in influencing audience members to be persuaded to one side or the other for each case (\textit{a priori} undecided or decided) of persuasion. Encoding audience-level and linguistic factors as features, we structure the prediction task as follows:
Given an individual voter, predict which debater/side (\textsc{PRO} or \textsc{CON}) the voter will be convinced by after the debate. We experiment with the features described in Section \ref{features}.

We consider only samples from the data where (1) a voter was undecided before the debate and then adopted a stance, i.e., \ voted for one of the debaters as the winner (\textsc{from-middle}); and (2) a voter was (seemingly) decided beforehand and then flipped their stance \textsc{from-opposing}. We do \textit{not} consider samples where (1) a voter declared a ``tie" between the debaters after the debate; and (2) a voter was decided beforehand and voted for the debater with the stance that they agreed with beforehand. To study the effect of each of the debaters' linguistic and user-based features on persuasion, we specifically look at which side (\textsc{PRO} vs.\ \textsc{CON}) did the convincing for a particular voter. Figure~\ref{votes-ex} illustrates example user votes for each of the two cases. Distinguishing instances of voters being persuaded into these case groupings allows us to examine what makes an argument persuasive to undecided versus decided audience members.  Table~\ref{data-stats} summarizes the dataset statistics relevant to the voter cases.

\begin{table}[t!]
\begin{center}
\begin{tabular}{|lcc|}
\hline \textbf{Persuasion Case} & \textbf{\# instances} & \textbf{\# debates} \\ \hline
\textsc{from-middle} & 4,360 & 3,652 \\
\textsc{from-opposing} & 2,642 & 2,183 \\
\hline
\end{tabular}
\end{center}
\caption{\label{data-stats} Number of voters in \textsc{from-middle} and \textsc{from-opposing} categories. }
\end{table}

\subsection{Differences Between Persuasion Groups}

We find distinct differences in the important features for predicting the outcome for voter groups \textsc{from-middle} and \textsc{from-opposing}. Best-performing set of linguistic features for \textsc{from-middle} includes all features minus the \textit{use of citations}, \textit{referring to the opponent}, and \textit{swear words}, while the best-performing set of linguistic features for \textsc{from-opposing} includes all features minus \textit{subjectivity}, \textit{modals}\footnote{The usage of modal verbs, i.e.,\ \textit{can}, \textit{should}, \textit{will}, and \textit{may}.}, and \textit{bi-/tri-gram TF-IDF}.\footnote{Calculated with a maximum of 30 terms.}

The set of linguistic features that are important for each the two groups have subtle differences in nature. A possible analysis that distinguishes the groups is that there is a difference in the rhetorical strategies that are the most effective. The use of modals, subjectivity, and general word choice are semantic features of an argument that can affect the perception of an argument's content. Based on our results, these content-based features are more important for undecided voters than for decided voters. In comparison, the use of swear words, citing sources, and referring to the opponent are stylistic features of an argument that can affect the perception of the debater. Our results indicate that these style-based features are not as important for undecided voters as for decided voters. This account is consistent with the findings of \citet{schill} that undecided voters respond most to content-rich rhetorical strategies and the findings of \citet{vecc,sween} that decided voters tend to selectively attend to information in a message based on prior attitudes. The account is also in line with experiments conducted by \citet{adams}, which found that affiliated voters do not adjust their positions in response to a party's actual policy statements but instead adjust their positions based on their subjective perceptions of the party. We have further found that audience-level aspects are comparatively more predictive of outcomes for undecided voters. 

\section{Limitations}
In this study, we develop a framework to account for users' prior beliefs in their opinion formation. We mainly focus on users' political and religious ideologies and whether they are undecided vs. decided a priori. However, there are many user aspects such as debating experience, prior interactions, education level, etc., which can impact their opinion formation. We do not propose a method to account for all these factors simultaneously. Moreover, we do not suggest any causal implications since our findings are correlational. 


\section{Chapter Summary}

In this chapter, we study the effect of the users' prior beliefs (i.e., political and religious ideology) and their initial stance on persuasion. We formulate the prediction task of determining which debater an individual voter finds persuasive in order to study the effect of these factors. We show that prior beliefs play a crucial role in this task. Furthermore, we explore the factors that govern persuasion for an a priori undecided vs. decided audience and find differences in the most predictive features for persuasion.

\chapter{Modeling the Effect of Social Interaction in Computational Persuasion} \label{social_interactions}

In Chapter \ref{prior_beliefs}, we study the impact of prior beliefs on persuasion. In this chapter, using the DDO dataset described in Chapter \ref{dataset}, we explore the effect of a user's social interaction on their debating success, considering all the debates that the user participates in over time. 

\section{Background}
There has been a tremendous amount of research on understanding user interactions and behaviour on social media \citep{backstrom2011center,nagarajan2010qualitative,macskassy2011people,Maia:2008:IUB:1435497.1435498,Benevenuto:2009:CUB:1644893.1644900,burke2009feed,golder2007rhythms,wilson2009user,lim2015mytweet, kumar2011understanding}. For example, \citet{wilson2009user} analyze the interaction graphs of Facebook user traces and show that interaction activity on
Facebook is significantly skewed towards a small portion
of each user's social links.
\citet{lim2015mytweet} investigates how people interact in multiple online social networks. It has been further shown that there is a strong relationship between a user's social interaction and their influence on social media. For example, \citet{10.1145/1963192.1963250} and \citet{cha2010measuring} and have shown that individuals with
more activity and personal engagement are more influential on Twitter.  Although there is a lot of work on understanding user behavior on social media sites such Facebook and Twitter, understanding the influence of user behavior on their persuasion success on debating platforms has been limited.

\citet{10.1145/1963192.1963250} is the most similar to our work, in that the authors study the effect of interaction dynamics, such as participant entry order and degree of back-and-forth exchange in the discussion, on success in changing an opinion holder's stance in a thread. Note that, unlike our study, this work does not consider the effect of social interaction features (such as friendship network or voter network) on users' \textit{success}. Moreover, we study the overall \textit{success} of users over their lifetime, rather than a single debate or discussion thread. 

We hypothesize that it is essential to account for the effect of social interactions in computational persuasion. Success in persuasion might also depend on an individual's social interaction and engagement with other users (on the debate platform) over time. For example, being more engaged with others over time may expose an individual to more diverse ideas and people, which could foster argumentation skills that are more applicable to convincing a more diverse audience. Focusing on only individual debates and discussion threads, prior work has not investigated the relative effect of an individual's social interaction, personal traits, and language
use on their success in persuasion. In this chapter,  we focus on online
debates and study success over a user's lifetime by looking at interaction and engagement with the community over time, rather
than focusing on individual debates to understand the relative impact of these factors on a user's success in persuasion. 

\section{Methodology} 

Our study employs the DDO (debate.org) dataset described in Section \ref{dataset}. Its extensive user information and multiple well-structured debates/interactions per user provides a unique opportunity to study users' success over time while accounting for the effect of individuals'
social interactions, personal traits, and language use. Users provide demographic information as well as their stance on controversial topics. They interact with one another in many ways: 1) debating, 2) evaluating the performance of other debaters, 3) commenting on debates, 4) asking/answering opinion questions, 5) voting in polls, 6) creating polls, 7) becoming friends. 

\subsection{Task Description}

This section describes the methods used to investigate the underlying dynamics of success in online debate. First, we explain how we measure the users' success, and then we explore the role of personal traits, social interactions, and language in predicting success. 

\subsection{User Success} \label{section_determining_success}

We compute the overall \textit{success} in debating for a user \textit{u} as:

\begin{equation} 
     \text{\textit{success}}_{u} = \frac{\text{number of debates \textit{u} won}}{\text{number of total debates \textit{u} participated in as a debater}}
\end{equation}

We treat users with  $\text{\textit{success}}_{u}\geq70$\% as \textit{successful}, $\text{\textit{success}}_{u}\leq30$\% as \textit{unsuccessful} and $30$\%$<\text{\textit{success}}_{u}<70$\% as \textit{mediocre}. 

 \subsection{Prediction Task} \label{user_interaction_prediction_task}

To understand the relative effect of a user's personal traits, social interaction, and language on their \textit{success}, we study the following prediction task: \textbf{given a pair of debaters where one of them is \textit{successful}, and the other is \textit{unsuccessful} over the second and third stage of their lifetime, predict the \textit{successful} one}. Note that while determining our label for success, we consider only the debates in the second and third stage of a user's lifetime to be able to study the relative effect of \textit{success} in their first life stage (\textit{success prior}) vs. other factors in a controlled way. We experiment with two settings where we control for the effect of \textit{debate experience} and \textit{success prior} respectively.

\textbf{{\sc setting 1}}. To \textbf{control the effect of debate experience} in \textit{success},  we create the pairs by \textbf{matching users according to the number of debates} that they participated in (i.e., users within a pair have the same number of debates).\footnote{There are 2,154 such pairs in our dataset.}

\textbf{{\sc setting 2}}. Given that we're interested in understanding the factors that correlate with \textit{success}, we \textbf{control for the \textit{success prior}} in a very specific way -- we only consider users that were \textit{unsuccessful} in their initial life stage (\textbf{success prior$\leq30\%$}\footnote{There are 957 such pairs in our dataset.}). This allows us to directly study the factors correlated with users that were initially \textit{unsuccessful}, but later went on to become \textit{successful} debaters. 

In the following subsections, we describe each of the factors (i.e., personal traits, social interactions, and language) that we study in our experiments. 

\subsubsection{Personal Traits}
In Chapter \ref{prior_beliefs}, we describe our findings on the role of prior beliefs on users' persuasion success in online argumentation, looking at the individual debates. We further investigate this effect in a debater's \textit{success} over their lifetime. We also extend this study by considering additional personal traits, such as the degree to which a debater's demographic (e.g., gender and ethnicity) matches those of their friends and the voters participating in the debates.

We extract features to encode the similarity for a user's opinion, political ideology, religious ideology, gender, and ethnicity with that of her friends and voters. To compute opinion similarity, we used the information about users' opinions on the \textit{big issues}.\footnote{We consider issues where users identified their side as either {\sc pro} or {\sc con} and measure the similarity of their opinion for these issues with their friends and voters.} 

\begin{figure*}[t]
\centering

\subfigure[Opinion Similarity]{%
{\epsfig{file = 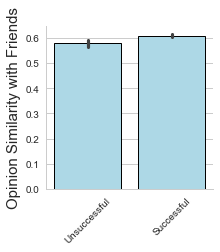, width = 6cm}}%
\label{fig:friends_sim}%
}\qquad
\subfigure[Religious Ideology Match]{%
{\epsfig{file = 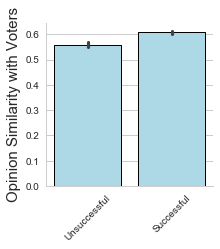, width = 6cm}}%
\label{fig:voter_sim}%
}
\caption{Similarity of Unsuccessful vs. Successsful Users with their Friends and Voters.}
\label{similarity_voters_friends}
\end{figure*}

Figure \ref{similarity_voters_friends} shows the similarity of \textit{successful} and \textit{unsuccessful} users' personal traits with that of their friends and voters respectively. We find that \textit{successful} users have significantly higher opinion similarity with their friends than \textit{unsuccessful} users. Moreover, they have significantly higher opinion similarity, religious ideology match, gender match, and ethnicity match with voters than \textit{unsuccessful} users. This implies that having voters with a similar background may be an important factor for \textit{success}, since an audience's decision about the performance of debaters may be influenced by the extent to which their prior beliefs match \citep{durmus-cardie-2018-exploring}. 

\subsubsection{Social Interaction}

The users interact with each other on the platform in the following ways: 1) debating 2) evaluating the performance of other debaters, 3) commenting on debates, 4) asking/answering opinion questions, 5) voting in polls, 6) creating polls, 7) becoming friends. We present examples for an opinion question, an opinion argument, and a poll topic below: 

\begin{quote}
    \textbf{Example Opinion Question.} "Does God exist?" \footnote{Full discussion on the topic can be found at \href{https://www.debate.org/opinions/does-god-exist}{https://www.debate.org/opinions/does-god-exist}.} \\ 
    \textbf{Example Opinion Argument.} "He probably does not exist. I don't think that it's possible to say yes or no either way. We can only conclude that there is more logical evidence to say that a God probably does not exist, ..." \\
    \textbf{Example Poll Topic.} Do you believe in Evolution or Creationism?  
\end{quote} 

We hypothesize that modeling these interactions is important to understand the differences between how \textit{successful} and \textit{unsuccessful} users interact on this platform and whether or not these are important factors for success. The ability to interact with others in a myriad of different ways provides users with ample opportunity to learn interesting new strategies and improve their skills over time, as they are exposed to a diverse set of perspectives.

\begin{figure*}
\centering
\subfigure[Number of votes]{%
{\epsfig{file = 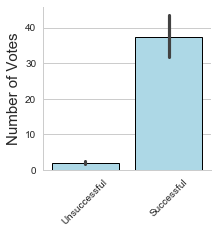, width = 5cm}}%
\label{fig:votes_activity}%
}\qquad
\subfigure[Number of opinion arguments]{%
{\epsfig{file = 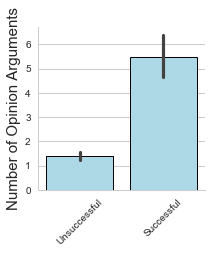, width = 5cm}}%
\label{fig:op_activity}%
}\qquad
\subfigure[Number of poll votes]{%
{\epsfig{file = 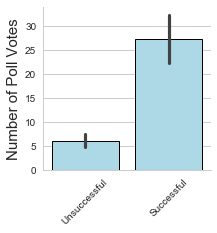, width = 5cm}}%
\label{fig:poll_activity}%
}\qquad
\subfigure[Number of friends]{%
{\epsfig{file = 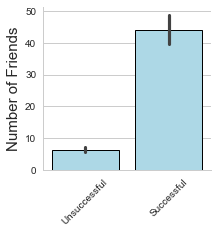, width = 5cm}}%
\label{fig:num_friends}%
}
\caption{Interaction statistics for  \textit{unsuccessful} and \textit{successful} users. \textit{Successful} users have significantly higher participation on the platform than \textit{unsuccessful} users.}
\label{interactions:activity_stats}
\end{figure*}

Figure \ref{interactions:activity_stats} shows the interaction statistics for \textit{successful} and \textit{unsuccessful} users.\footnote{We controlled for number of debates to remove the effect of ``being a new user'' by pairing \textit{successful} and \textit{unsuccessful} users according the number of debates they participated in.} We see that, overall, \textit{successful} users have significantly higher participation on the platform. 

\textbf{Friendship network.} 
We represent the friendship network as an \textbf{undirected} graph $G = (V, E)$ where $V$ represents the set of users, and $E$ represents the set of edges where $(x,y)\in E$ if $x \in V$ and $y \in V$ are friends.  

\textbf{Voter network.}
We represent the voter network as a weighted \textbf{directed} graph $G = (V, E)$ where $V$ represents the set of users, and $E$ represents the set of edges where $(x,y)\in E$ if $x \in V$ voted in a debate in which $y \in V$ participated as a debater. The weight of the graph represents how many times $x$ voted in debates $y$ was a debater. Note having $(x,y)$ edge in the graph \textbf{does not} imply that $x$ voted for $y$ in a debate. 

\begin{figure*}
\centering
\subfigure[Hub Score - Voter Network ]{%
{\epsfig{file = 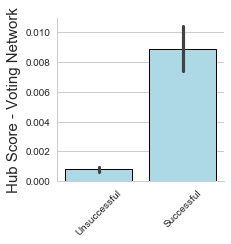, width = 5cm}}%
\label{fig:hub_score}%
}\qquad
\subfigure[In-degree Centrality - Voter Network]{%
{\epsfig{file = 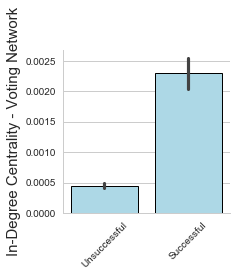, width = 5cm}}%
\label{fig:in_degree_cent}%
}\qquad
\subfigure[Out-degree Centrality - Voter Network]{%
{\epsfig{file = 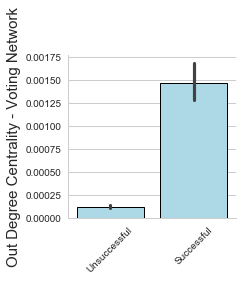, width = 5cm}}%
\label{fig:out_degree_cent}%
}\qquad
\subfigure[Centrality - Friendship Network]{%
{\epsfig{file = 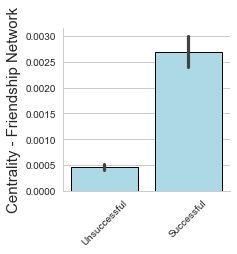, width = 5cm}}%
\label{fig:centrality}%
}\qquad
\subfigure[Page Rank - Friendship Network]{%
{\epsfig{file = 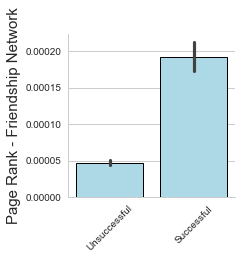, width = 5cm}}%
\label{fig:page_rank}%
}
\caption{Characteristics of voter and friendship network for \textit{successful} and \textit{unsuccessful} users.}
\label{interactions:graph_char}
\end{figure*}

\textbf{Hubs and authorities in voter network.}
Using the HITS algorithm \citep{kleinberg1999authoritative}, we compute hub and authority scores for each node (user) in the voter network graph.
We expect that users that participate in debates as debaters are the authoritative sources of information on the controversial topics on this platform; therefore, they should have higher authority scores. On the other hand, users with higher hub scores represent people who may not necessarily be authoritative sources of information on the topic, but they are interested in the topic and; therefore, by providing feedback, they lead other users to these debates.  
We find that \textit{successful} users have, on average, a significantly higher hub score than \textit{unsuccessful} users (p $<$ $0.001$). As shown in Figure \ref{interactions:graph_char}, we further observe that 
\textit{successful} users have, on average, a  significantly higher in-degree centrality and out-degree centrality than \textit{unsuccessful} users in the voter network. Similarly, \textit{successful} users have higher degree centrality and page rank  than \textit{unsuccessful} users in their friendship network. 

\subsubsection{Language}
To capture the linguistic style of the debaters' language and its relationship to their \textit{success}, we use textual features that encode 1) users' own language and 2) the interplay between users' and their opponents' language.

\begin{table*}[ht]
\centering
\begin{tabular}{|p{4cm}|p{10cm}|}
\hline
Aspect &  Features\\
\hline
\textbf{Personal Traits} & 1) \textbf{match of the personal traits (e.g., gender, political ideology, religious ideology and ethnicity) with friends and voters}. \\ 
 
& 2) \textbf{opinion similarity with friends and voters}.\\ 
\hline
\textbf{Social Interactions} & 1) \textbf{participation features} : \# of comments, \# of votes, \# of friends, \# of opinion questions and arguments, \# of voted debates, \# of poll votes and topics. \\
& 2) \textbf{friendship network features} : degree, degree centrality, page rank scores. \\
& 3) \textbf{voter network features}: in-degree, out-degree, in-degree centrality, out-degree centrality, page rank, hub and authority scores.
\\
\hline
\textbf{Language} & 1) \textbf{features of debaters' own language} : \# of words, \# of definite articles, \# of  indefinite articles, \# of  person pronouns, \# of positive words, \# of negative words, \# of hedges, \# of swear words, \# of punctuation, \# of links, average sentiment, type-token ratio, \# of quotes, distribution of POS tags, distribution of named entities, BOW.\\
& 2) \textbf{features to encode the interplay} : exact content word match, exact stop word match, content word match with synonyms.\\
\hline
\end{tabular}
\caption{Personal Traits, Social Interactions and Language Features.}
\label{features}
\end{table*} 

\textbf{Modeling users' own language.}
We extract features from the text of users' debates, opinion questions, opinion arguments, poll votes, and poll topics. These features include \# of words, word category features (e.g., \# of personal pronouns, \# of positive and negative words), structural features (e.g., distribution of POS tags and named entities), and features to encode the characteristics of the entire language (e.g., type-token ratio)

\textbf{Modeling interplay between a debater and their opponent.}
We measure the interplay between debaters and their opponents by measuring how similar a debater's language is to the previous statement made by her opponent. To measure the similarity of a debater's language (D) to that of the opponent's (O) in a round, we look at \# of content words that are in both D and O, \# of stop words that are in both D and O and \# of content words that are in D and have synonyms in O. 

The \textit{content word match with synonyms} feature aims to capture the cases where the opponent refers to similar concepts but does not necessarily use the same words as the debater. 

The complete list of features modeling the aspects of personal traits, social interactions, and language features is shown in Table \ref{features}.

\subsection{Prediction Results} \label{results}
We use weighted logistic regression and choose the amount and
type of regularization ($\ell$1 or $\ell$2) by grid search over five cross-validation
folds. We compute \textbf{weighted} precision, recall and F1 scores.

In {\sc setting 1}, we create user pairs ($u_1$,$u_2$) where:
\begin{itemize}
    \item $u_1$ and $u_2$ have an equal number of debates they participated in as debaters.
    \item One of $u_1$ or $u_2$ is \textit{successful} and the other one is \textit{unsuccessful} over the second and third stage of their lifetime.\footnote{We consider success only over the second and third stage of users' lifetime in our prediction task, in order to study the effect of \textit{success prior} vs. the other aspects. We use the success in the first life stage as \textit{success prior}.}
\end{itemize} 

In {\sc setting 2}, in addition to the requirements of {\sc setting 1}, we also require $u_1$ and $u_2$  to both have \textit{success prior} $\leq$ 0.3.

\textbf{Task.} For both {\sc setting 1} and {\sc setting 2}, we aim to predict whether $u_1$ or $u_2$ is \textit{successful} over the second and third stage of her lifetime.

In {\sc setting 2}, by only studying user pairs with low \textit{success priors}, we aim to understand the factors that are important for a user to improve as a debater over time.

\begin{table*}
\begin{center}
\begin{tabular}{|p{3.4cm}|p{4.9cm}|p{1.6cm}|p{1.6cm}|p{1.6cm}|}

    \hline
    
    & Feature  & Precision & Recall & F1 \\ 
    
    \hline

    & (1) Majority &   $26.47_{\pm1.11}$   &  $51.44_{\pm1.08}$ & $34.95_{\pm1.22}$ \\
    
    & (2) Debating experience &   $52.70_{\pm2.91}$   &  $52.04_{\pm1.77}$ & $41.76_{\pm2.06}$  \\

    & (3) Success prior &   $65.20_{\pm0.77}$   &  $64.39_{\pm0.65}$ & $63.63_{\pm0.50}$  \\ 
    
    \hline
    Personal Traits &
   
    (4) Overall similarity with voters &  $61.93_{\pm1.60}$   &  $60.86_{\pm1.70}$ & $59.44_{\pm1.67}$ \\ 
  
    & (5) Overall similarity with friends &  $62.70_{\pm0.86}$   &  $59.98_{\pm1.05}$ & $56.94_{\pm1.14}$ \\

    \hline

    Social Interaction  &
    (6) Participation features & $67.78_{\pm1.66}$   &  $66.02_{\pm2.33}$ & $64.82_{\pm2.70}$ \\
    
    & (7) Friendship network features & $64.23_{\pm1.40}$   &  $63.60_{\pm1.40}$ & $62.92_{\pm1.35}$ \\
    
    & (8) Voter network features & $72.39_{\pm0.19}$   &  $70.75_{\pm0.34}$ & $70.20_{\pm0.70}$ \\
    
    & (6) + (7) + (8) & $72.67_{\pm0.73}$   &  $72.29_{\pm0.93}$ & $72.12_{\pm1.03}$ \\ 
   
    \hline
    
    Language &
    (9) \# of words & $70.37_{\pm1.41}$ & $70.15_{\pm1.55}$ &  $69.97_{\pm1.59}$  \\ 
    
    & (10) Features of debaters' interplay &  ${62.11}_{\pm{1.09}}$ &  ${62.07}_{\pm{1.03}}$ &  ${61.92}_{\pm{1.01}}$ \\
    
    & (11) Features of debaters' own language &  ${72.65}_{\pm{2.45}}$ &  ${72.66}_{\pm{2.45}}$ &  ${72.64}_{\pm{2.44}}$ \\ 
    \hline
 
    {Combinations}&
  (6) + (7) + (8) + (11) & $78.49_{\pm1.29}$ & $78.46_{\pm1.32}$ &  $78.45_{\pm1.32}$ \\ 
   
  & (6) + (7) + (8) + (10) + (11) & $\mathbf{81.63}_{\pm\mathbf{1.63}}$ & $\mathbf{81.62}_{\pm\mathbf{1.65}}$ & 
  $\mathbf{81.61}_{\pm\mathbf{1.65}}$  \\ 

\hline
\end{tabular}
\end{center}
\caption{Prediction Task Results for {\sc setting 1}. Voter network features are the most predictive social interaction features. Combining interaction and language features achieves the best predictive performance.}

\label{tab:results_setting_1}
\end{table*}

\subsubsection{Results for {\sc setting 1}}
Table \ref{tab:results_setting_1} shows the results for {\sc setting 1}. We compare our model with three simple baselines -- majority, debating experience, and success prior. For the majority baseline, we predict the most common label in the training data for each test example. For debating experience baseline, we use \# of debates as the only feature to predict the \textit{successful} debater. For success prior baseline, we pick the user with the higher \textit{success prior} as successful. 

In {\sc setting 1}, since we do not control for the \textit{success} in the first life stage, we see that the \textit{success prior} information alone can achieve $63.63\%$ F1 score. This implies that there is a correlation between users' \textit{success} in their early life stage and later life stages. This factor may be related to users' prior debating skills. We observe that the features that encode debaters' overall similarity with voters and friends achieve significantly better F1 scores than majority and debating experience baselines. However, these features do not have as high a predictive power as the \textit{success prior}. We perform an ablation study for participation features, friendship network features, and voter network features. We find that voter network features are significantly more predictive than the baselines, personal trait features, and other social interaction features. We also perform an ablation study for the language features and find that \# of words is a very predictive feature of \textit{success}. When we combine the language features with the interaction features, we get the best predictive performance (81.61\% F1 score) for this task which is significantly better than the baselines. This indicates that it is important to account for social interaction and language factors to determine the \textit{successful} debater since these two components encode different kinds of information about the users.

\begin{table*}
\begin{center}
\begin{tabular}{|p{3.4cm}|p{4.9cm}|p{1.6cm}|p{1.6cm}|p{1.6cm}|}

    \hline
    
    & Feature  & Precision & Recall & F1 \\ 
    
    \hline

    & (1) Majority &   $26.67_{\pm1.61}$   &  $51.62_{\pm1.56}$ & $35.16_{\pm1.76}$ \\
    
    & (2) Debating experience &  $46.00_{\pm0.89}$  &  $50.16_{\pm1.02}$ & $38.98_{\pm3.92}$  \\

    & (3) Success prior &  $55.60_{\pm0.93}$   &  $55.07_{\pm0.39}$ & $52.10_{\pm0.47}$ \\ 
    
    \hline
    Personal Traits &
   
    (4) Overall similarity with voters &  $56.55_{\pm2.43}$   &  $55.69_{\pm1.31}$ & $52.68_{\pm1.47}$ \\ 
  
    & (5) Overall similarity with friends & $55.87_{\pm3.43}$   &  $54.23_{\pm2.35}$ & $47.52_{\pm3.18}$ \\

    \hline

    Social Interactions  &
    (6) Participation features & $59.39_{\pm4.09}$   &  $57.68_{\pm2.34}$ & $55.08_{\pm3.16}$\\
    
    & (7) Friendship network features & $57.94_{\pm1.87}$   &  $57.16_{\pm1.50}$ & $55.41_{\pm1.67}$ \\
    
    & (8) Voter network features & $70.54_{\pm1.78}$   &  $69.91_{\pm1.79}$ & $69.65_{\pm1.76}$\\
    
    & (6) + (7) + (8) & $71.66_{\pm0.71}$   &  $71.47_{\pm0.51}$ & $71.38_{\pm0.51}$ \\ 
   
    \hline
    
    Language &
    (9) \# of words & $65.78_{\pm0.85}$ & $64.99_{\pm1.03}$ &  $64.41_{\pm1.16}$  \\ 
    
    & (10) Features of debaters' interplay &  ${57.47}_{\pm{1.42}}$ &  ${57.16}_{\pm{1.31}}$ &  ${56.41}_{\pm{1.29}}$ \\
    
    & (11) Features of debaters' own language &  ${64.48}_{\pm{0.74}}$ &  ${64.37}_{\pm{0.90}}$ &  ${64.24}_{\pm{0.97}}$ \\ 
    \hline
 
    {Combinations}&
  (6) + (7) + (8) + (11) & $75.44_{\pm0.90}$ & $75.44_{\pm0.90}$ &  $75.43_{\pm0.89}$ \\ 
   
  & (6) + (7) + (8) + (10) + (11) & $\mathbf{78.06}_{\pm\mathbf{0.88}}$ & $\mathbf{78.05}_{\pm\mathbf{0.89}}$ & 
  $\mathbf{78.05}_{\pm\mathbf{0.88}}$ \\ 

\hline
\end{tabular}
\end{center}
\caption{Prediction Task Results for {\sc setting 2}. Similar to {\sc setting 1}, voter network features are the most predictive social interaction features, and combining interaction and language features achieves the best predictive performance.}

\label{tab:results_setting_2}
\end{table*}

\subsubsection{Results for {\sc setting 2}}
In this task, by controlling for \textit{prior success}, we aim to understand the factors correlated with \textit{success} by reducing the effect of prior debating skills of the users.  
As shown in Table \ref{tab:results_setting_2}, the F1 score for the \textit{success prior} baseline is not as quite as high as in {\sc setting 1}, since we control for this aspect by ensuring both users in the pair are \textit{unsuccessful} in their initial life stage. However, this does not necessarily mean that the two paired users will have the same success prior, which explains why success prior still performs better than the other baselines.
We do not observe any significant difference between the performance of the features encoding personal traits, participation, and the baseline. However, consistent with the {\sc setting 1}, we see that features of the voter network are significantly better (69.65\%) in predicting \textit{success}. 
Although language features achieve a significantly better F1 score than the baseline, they perform significantly worse than the voter network features. Similar to {\sc setting 1}, combining these language features with the social interaction features improves the performance significantly (78.05\% F1 score).

\subsubsection{Feature Analysis}

To understand the important social interaction and language features, we 1) compute the correlation coefficients for the feature values and the labels, 2) analyze the coefficients of the logistic regression classifier, and 3) apply the recursive feature elimination method \citep{guyon2002gene} to rank features according to their importance. In this section, we present the consistently important features for each of these methods. 

\textbf{Analysis of Social Interaction Features.}
We find that the \textbf{most important} social interaction features for {\sc setting 1} are {authority score}, {hub score}, {in and out-degree centrality} and the {page rank} of the voter network. Note that all these important features are \textbf{positively correlated} with \textit{success}. Although participation and friendship network features (e.g., \# of voted debates, degree of the user node in friendship network) are also positively correlated with \textit{success}, the correlation values for these are not as high as the ones of the voter network features. We also find a high correlation between some of the user activities. For example, users with more \# of comments are more active in making friends, voting, providing poll votes, and having higher centrality value in the friendship network. Perhaps surprisingly, we do not observe any correlation between \# of voted debates and hub/authority scores in the voter network. However, we see a highly positive correlation between hub scores, authority scores, in-degree centrality, out-degree centrality, and page rank values of the voter network. This implies that \textit{success} is not only about the quantity of voted debates but also about the characteristics of the debaters involved in these debates, since the hub score of a user is influenced by the authority scores of the debaters they vote for. Similarly, the authority score of a user is influenced by the hub scores of the voters that participate in her debates. Therefore, besides the frequency of interaction, the type of the interaction and characteristics of users involved in the interaction are important to consider. 
Consistent with {\sc setting 1}, in {\sc setting 2}, the most important features (positively correlated with \textit{success}) are 
{authority score}, {hub score}, {in and out-degree centrality} and the {page rank} of the voter network. We observe the same patterns of user activities and authority and hub scores as in {\sc setting 1}. 

\textbf{Analysis of Language Features.}
We find that number of words is positively correlated with \textit{success}. It may be the case that longer text may convey more information and explain the points more explicitly \citep{doi:10.1080/00028533.1997.11978023,doi:10.1080/00028533.1998.11951621}. The bag of words feature is not as predictive as the \# of words feature.
For both {\sc setting 1} and {\sc setting 2}, we observe that the value of average {sentiment} is negatively correlated with \textit{success}. The reason for this may be that negative information is more attention grabbing than positive information \citep{article_ditto,doi:10.1080/00913367.1992.10673357,article_pratto} since people are more used to seeing arguments that are phrased in a more positive way \citep{meyerowitz1987effect}. 
We also find that \textit{type-token ratio} (diversity of language) is negatively correlated with \textit{success} for both settings. It may be the case that people who talk about a smaller set of topics gain expertise on these topics over time; therefore, they may be more \textit{successful}. 
We observe that other textual features are positively correlated with \textit{success} for both of these settings. However, the degree of correlation is not as high as it is for {type-token} ratio and {sentiment}.

\begin{table*}[ht]
\hspace{-0.5cm}
\begin{center}
 \begin{tabular}{|p{3.5cm}|p{4.9cm}|p{1.6cm}|p{1.6cm}|p{1.6cm}|p{1.6cm}|}
 \hline

    & Feature  & Precision & Recall & F1  \\ 
    \hline
    \multirow{3}{*}{}
    &(1) Majority &   $26.97_{\pm2.69}$   &  $51.86_{\pm2.62}$ & $35.46_{\pm2.95}$ \\&
    
    (2) Debating experience &   $53.77_{\pm2.95}$   &  $52.43_{\pm2.91}$ & $43.02_{\pm6.19}$ \\&
    
    (3) Success prior &   $39.94_{\pm7.63}$   &  $51.00_{\pm2.23}$ & $36.04_{\pm2.35}$\\
    \hline
    
    \multirow{3}{*}{Personal Traits} &
   
    (4) Overall similarity with voters &  $55.17_{\pm1.58}$   &  $55.00_{\pm2.36}$ & $53.94_{\pm2.99}$ \\&
  
    (5) Overall similarity with friends &  $66.38_{\pm4.11}$   &  $63.43_{\pm2.77}$ & $60.87_{\pm3.33}$\\

    \hline
    \multirow{5}{*}{Social Interactions}&
    (6) Participation features & $68.88_{\pm3.57}$   &  $68.00_{\pm2.86}$ & $67.88_{\pm2.96}$\\&
    (7) Friendship network features & $65.60_{\pm4.83}$   &  $64.00_{\pm3.81}$ & $62.81_{\pm3.73}$ \\&
    (8) voter network features & $64.36_{\pm1.57}$   &  $62.72_{\pm2.37}$ & $61.44_{\pm2.87}$\\&
    (6) + (7) + (8)& $67.80_{\pm1.86}$   &  $67.14_{\pm1.43}$ &  $66.97_{\pm1.42}$\\
   
    \hline
    \multirow{2}{*}{Language}&
    (9) \# of words & $67.63_{\pm3.90}$ & $66.57_{\pm2.70}$ &  $66.29_{\pm2.39}$ \\&
    (10) Features of debaters' interplay&  ${58.76}_{\pm{2.03}}$ &  ${57.43}_{\pm{0.86}}$ &  ${56.60}_{\pm{0.93}}$ \\&
    (11) Features of debaters' own language&  ${68.47}_{\pm{0.21}}$ &  ${68.14}_{\pm{0.14}}$ &  ${68.10}_{\pm{0.17}}$\\
    \hline
 
 \multirow{2}{*}{Combinations}&
 (6) + (7) + (8) + (11) & $69.32_{\pm2.48}$ & $69.00_{\pm2.42}$ &  $69.00_{\pm2.41}$\\ &
  (6) + (7) + (8) + (10) + (11) & $\mathbf{73.60}_{\pm\mathbf{0.80}}$ & $\mathbf{73.43}_{\pm\mathbf{0.70}}$ & 
  $\mathbf{73.43}_{\pm\mathbf{0.72}}$\\ 
\hline
\end{tabular}
\end{center}
\caption{Prediction Task Results for loss of \textit{success}. Participation features are the most important social interaction features. Combining the social interaction features with the language features gives the best prediction performance.}
\label{tab:results_failure}
\end{table*}

\section{Understanding the loss of \textit{success}}
In the previous section, we show that social interaction and language features are important to predict \textit{successful} debaters. Our findings are consistent for the case when 1) we only control for users' debating experience and 2) we also control for users' \textit{success prior}. Users' participation, the types of interactions they have on the platform, and the characteristics of the users they interact with are predictive of their \textit{success}, regardless of their prior expertise in debating (encoded by the \textit{success prior}).

In {\sc setting 1}, since we did not control for the \textit{success prior}, we studied the factors that are important for a user to become \textit{successful} in their second and third life stages, regardless of their \textit{success} in the beginning. In {\sc setting 2}, we studied the factors that are important for \textit{unsuccessful} users to improve their performance and become \textit{successful} over time. As a natural follow-up, we would also like to understand what factors are correlated with users who are initially \textit{successful}, but later become \textit{unsuccessful} in their lifetime.
 To do that, in {\sc setting 3}, in addition to the requirements of {\sc setting 1}, we have an additional criterion for all user pairs ($u_1$,$u_2$):
 \begin{itemize}
    \item $u_1$ and $u_2$ both have \textit{success prior} $\geq$ 0.7.\footnote{We have $700$ user pairs with these criteria.} 
\end{itemize}

\subsection{Results}
As shown in Table \ref{tab:results_failure}, features of personality traits, social interactions, and language perform significantly better than the baselines. For this task, the \textit{success prior} baseline performs relatively worse than in the previous two settings.
Upon closer examination, we observed that the variance of success priors for this task is an order of magnitude smaller than in {\sc setting 2}. Therefore, as a possible explanation, the \textit{success prior} may not be as predictive for this task.

In social interaction features, {similarity with friends} is the most predictive feature. However, {participation features} perform significantly better than the features of personal traits. For this task, contrary to {\sc setting 1} and {\sc setting 2}, we see that participation features are the most predictive in the set of social interaction features. This implies that a user's participation is important for them to remain \textit{successful}. Lower participation could be a contributing factor for these users to become unsuccessful eventually. Although friendship and voter network features are still significantly more predictive than the baselines, they are not as highly predictive as the participation features. For users with high \textit{success priors}, continued participation may be the most important aspect of their social interaction. 
We observe that language features alone achieve a similar performance as the social interaction features. Consistent with the {\sc setting 1} and {\sc setting 2}, combining social interaction and language features gives the best predictive performance (73.43\% F1 score).

\textbf{Analysis of Social Interaction Features.}
The \textbf{most important} social interaction features include \# of voted debates, degree of the user node in the friendship network, and hub
scores, authority scores, in-degree centrality, out-degree centrality and page rank values of voter network. All these features indicate higher participation on the platform, and they are positively correlated with staying \textit{successful}. Although the other social interaction features, such as {authority} and {hub} scores of the voter network are also positively correlated with \textit{success}, the value of correlation for these is not as high as the previously mentioned features. For users who are initially \textit{unsuccessful}, participation alone may not be enough for them to become \textit{successful} debaters -- the types of interactions and the characteristics of people with whom they interact are crucially important for their \textit{success}. On the other hand, users who are initially \textit{successful} may already be experienced debaters, and staying active and participating may be sufficient for them to remain \textit{successful}.

\textbf{Analysis of Language Features.}
As in {\sc setting 1} and {\sc setting 2}, \# of words is positively correlated with staying \textit{successful}.
We find that the {\# of first person pronouns} is the language feature with the highest positive correlation with staying \textit{successful}. We observe that users who refer to their personal experiences and opinions use {first person pronouns} more often. It may be the case that debaters may try to appeal to logos by citing personal experience \citep{cooper1992power}.
Consistent with {\sc setting 1} and {\sc setting 2}, the value of {average sentiment} is negatively correlated with staying \textit{successful}.

\section{Limitations}
In this study, we investigate the impact of social interaction on debating success. We find that higher participation and engagement improves the success of debaters over time. One potential reason is that users develop strategies to improve their debating skills. Another factor could be that users only participate in the topics they are comfortable with and do not improve their debating skills overall. Moreover, they may be debating with users that they are confident about defeating to increase their chances of winning. Therefore, in this setup, becoming more successful over time may not necessarily imply developing better argumentative skills. In future work, we would like to explore the effect of debate topics on users' success. Moreover, we aim to understand what characteristics of a user's language change over time and how it affects debating success.

\section{Chapter Summary}

This chapter explores the effect of a user's social interaction on their success in debating over time. We investigate the impact of language, personal traits, and social interaction simultaneously for predicting the successful debater given a pair of debaters where one of them is successful and the other is unsuccessful. We observe that successful debaters are significantly more engaged with others and more active on the platform. We find that a user's social interaction characteristics play a crucial role in determining their success in debates. We achieve the best predictive performance by combining social interaction features with features that encode information on language use.

\chapter{Modeling Pragmatic Context in Argument Impact Prediction}\label{context}

In the previous chapters, we discuss the impact of prior beliefs (Chapter \ref{prior_beliefs}) and social interactions (Chapter \ref{social_interactions}) on determining the more successful debater and a user's debating success over time, respectively. This chapter introduces a new dataset for argument impact prediction and explores methods to incorporate pragmatic context in determining argument impact. 

\section{Background}
Previous work in the social sciences and psychology has shown that the impact and
persuasive power of an argument depend not only on the language employed but also
on the credibility and character of the communicator (i.e., \ ethos)  \citep{fb566a52435647fcbb369ed48db6fbec, 1980-32482-00119790801, source-effect}, the traits and prior beliefs of the audience \citep{lord1979biased, edsfra.236698719980101, correll2004affirmed, hullett2005impact},
and the pragmatic context in which the argument is presented (i.e.,\ kairos) \citep{10.1086/209393, context_joyce}. 

Research in Natural Language Processing (NLP) has only
partially corroborated these findings. One very influential line of work, 
for example, develops computational methods to automatically determine
the linguistic characteristics of persuasive
arguments \citep{habernal-gurevych-2016-makes, 10.1145/2872427.2883081, zhang2016conversational}, but it does so without controlling for
the audience, the communicator, or the pragmatic context. Very recent work, on the other hand, shows that attributes of both the 
audience and the communicator constitute important cues for determining 
argument strength \citep{lukin2017argument,durmus-cardie-2018-exploring}.
They further show that audience and communicator attributes can influence the relative importance of linguistic features for predicting the persuasiveness 
of an argument. These results confirm previous findings in the social sciences that show a person's
perception of an argument can be influenced by their background and
personality traits.  
To the best of our knowledge, however, no NLP studies explicitly investigate the
role of \textit{kairos} --- a component of pragmatic context that refers to
the context-dependent ``timeliness" and ``appropriateness" of an argument and 
its claims within an argumentative discourse ---  in argument quality prediction.  

Among the many social science studies of attitude change, the order in which
argumentative claims are shared with the audience has been studied extensively:
\citet{10.1086/209393}, for example, summarize studies showing that the
argument-related claims a person is exposed to beforehand can affect his perception
of an alternative argument in complex ways.
\citet{context_joyce} similarly finds that changes in an argument's context can have a big impact on the audience's perception of the argument. 

Some recent studies in NLP have investigated the effect of interactions on the overall persuasive power of posts in social media \citep{10.1145/2872427.2883081,DBLP:conf/aaai/HideyM18}. However, in social media, not all posts have to express arguments or stay on topic \citep{DBLP:journals/corr/abs-1709-03167}, and qualitative evaluation of the posts can be influenced by many other factors such as interaction between the individuals \citep{Durmus:2019:MFU:3308558.3313676}. Therefore, it is difficult to measure the effect of argumentative pragmatic context alone in argument quality prediction without these confounding factors using the datasets and models presented in prior work.

In this chapter, we study the role of kairos on argument quality prediction by
examining the individual claims of an argument for their timeliness and appropriateness in the context of a particular line of argument. We define \textbf{kairos} as the sequence of \textbf{argumentative} text (e.g., claims) along a particular line of argumentative reasoning.
We first present a dataset extracted from \textit{kialo.com} of over 47,000
claims that are part of a diverse collection of arguments on 741 controversial
topics. The website's structure dictates that each argument must present a supporting or opposing claim for its parent claim, and stay within the topic of the main thesis. Rather than being posts on a social media platform, these are community-curated claims. Furthermore, for each presented claim, the audience votes on its impact within the given line of reasoning. Critically then, the dataset includes the argument context for each claim, allowing us to
investigate the characteristics associated with impactful arguments.

With the dataset in hand, we then propose the task of studying the characteristics of impactful claims by (1) taking the argument context into account, (2) studying the extent to which this context is important, and (3) determining the representation of context that is more effective. To the best of our knowledge, ours is the first dataset that includes claims with both impact votes and the corresponding context of the argument.

\section{Dataset}
\textbf{Claims and impact votes.} We collected claims from \textit{kialo.com}\footnote{The data is collected from this website in accordance with the terms and conditions.}\footnote{There is prior work by \citet{durmus-etal-2019-determining} which created a dataset of argument trees from \textit{kialo.com}. That dataset, however, does not include any impact labels.} for 741 controversial topics and their corresponding impact votes. The users of the platform provide impact votes to evaluate how impactful a particular claim is. Users can pick one of $5$ possible impact labels for a particular claim: {\sc no impact}, {\sc low impact}, {\sc medium impact}, {\sc high impact} and {\sc very high impact}. While evaluating the impact of a claim, users have access to the full argument context. Therefore, they can assess how impactful a claim is in the given context of an argument. Interestingly, in this dataset, the same claim can have different impact labels depending on the context in which it occurs.

\begin{figure*}[ht]
\centering
\includegraphics[width=15.9cm,height=9.7cm]{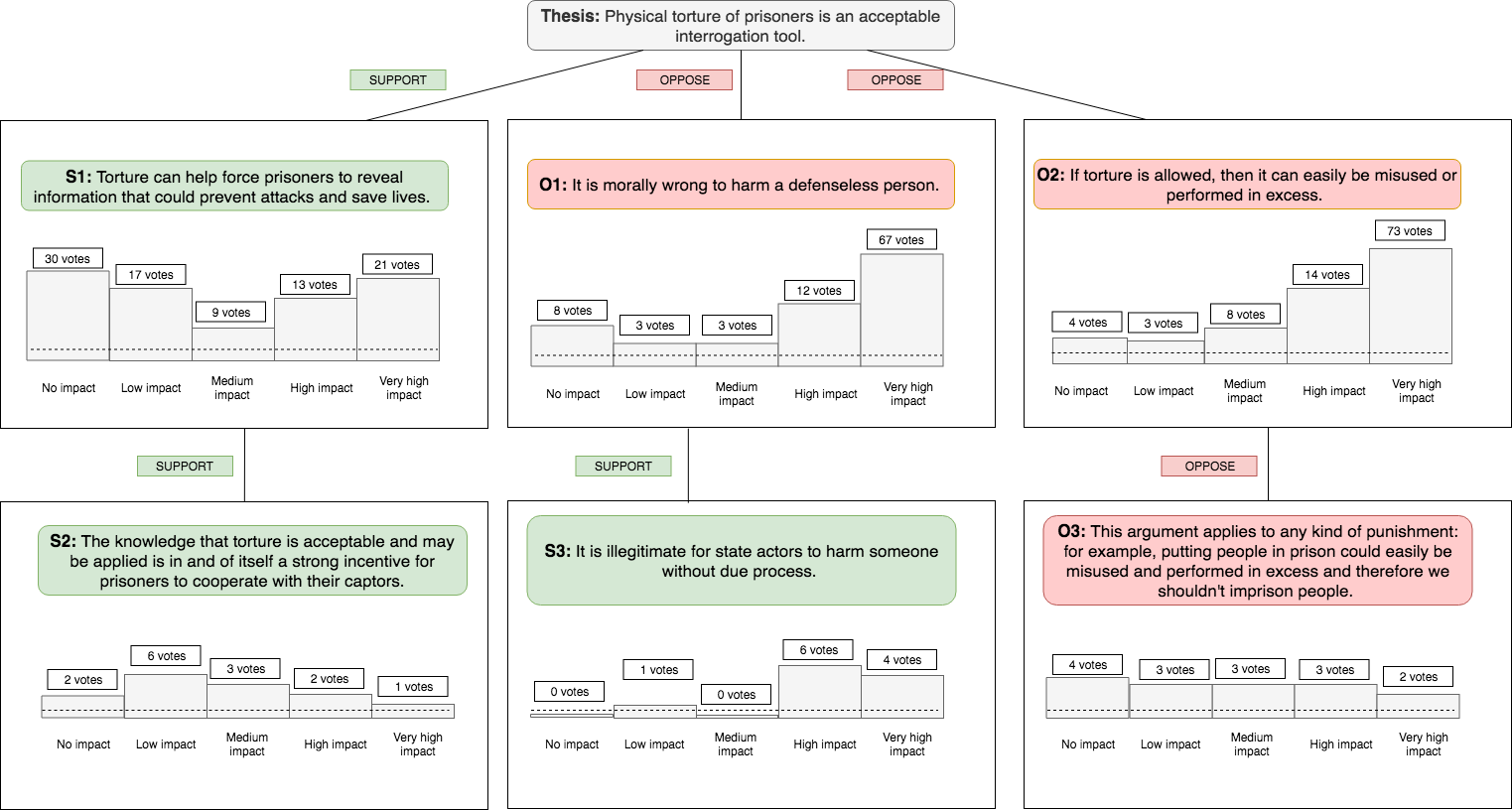}
\caption{Example partial argument tree with claims and corresponding impact votes for the thesis {\sc``Physical torture of prisoners is an acceptable interrogation tool.''}.}
\label{fig:impact_image}
\end{figure*}   

Figure \ref{fig:impact_image} shows a partial\textbf{ argument tree} for the argument \textbf{thesis} {\sc``Physical torture of prisoners is an acceptable interrogation tool.''}. Each node in the argument tree corresponds to a claim, and these argument trees are constructed and edited collaboratively by the users of the platform. 

Except for the thesis, every claim in the argument tree either opposes or supports its parent claim. Each path from the root to a leaf node corresponds to an \textbf{argument path} which represents a particular line of reasoning on the given controversial topic. 

\begin{figure*}
\centering
\subfigure[Number of trees with given range of total number of claims.] 
{\includegraphics[scale=0.7]{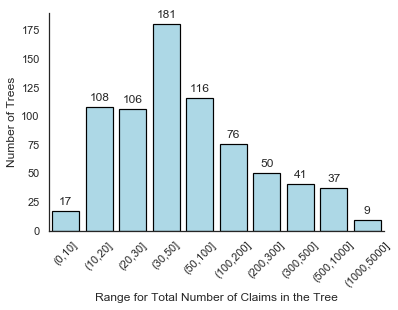} \label{fig:num_node}}\quad
\subfigure[Number of trees with given range of depth.]
{\includegraphics[scale=0.7]{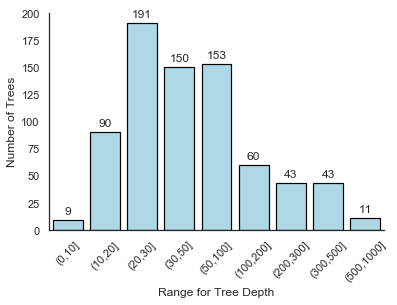} \label{fig:depth}}
\caption{Data statistics: For the majority of trees, the depth of the argument tree is $4$ or higher, and the argument tree has more than $30$ claims in the tree. Average number of claims and depth per argument tree are 127 and 5 respectively.}
\label{fig:data_stats}
\end{figure*}

The distribution of argument trees for a given range of claims, and depth is shown in Figures \ref{fig:num_node} and \ref{fig:depth} respectively. We see that for the majority of trees, the depth is $4$ or higher, and the number of claims is greater than $30$. 

\begin{figure}
\centering
\includegraphics[scale=0.75]{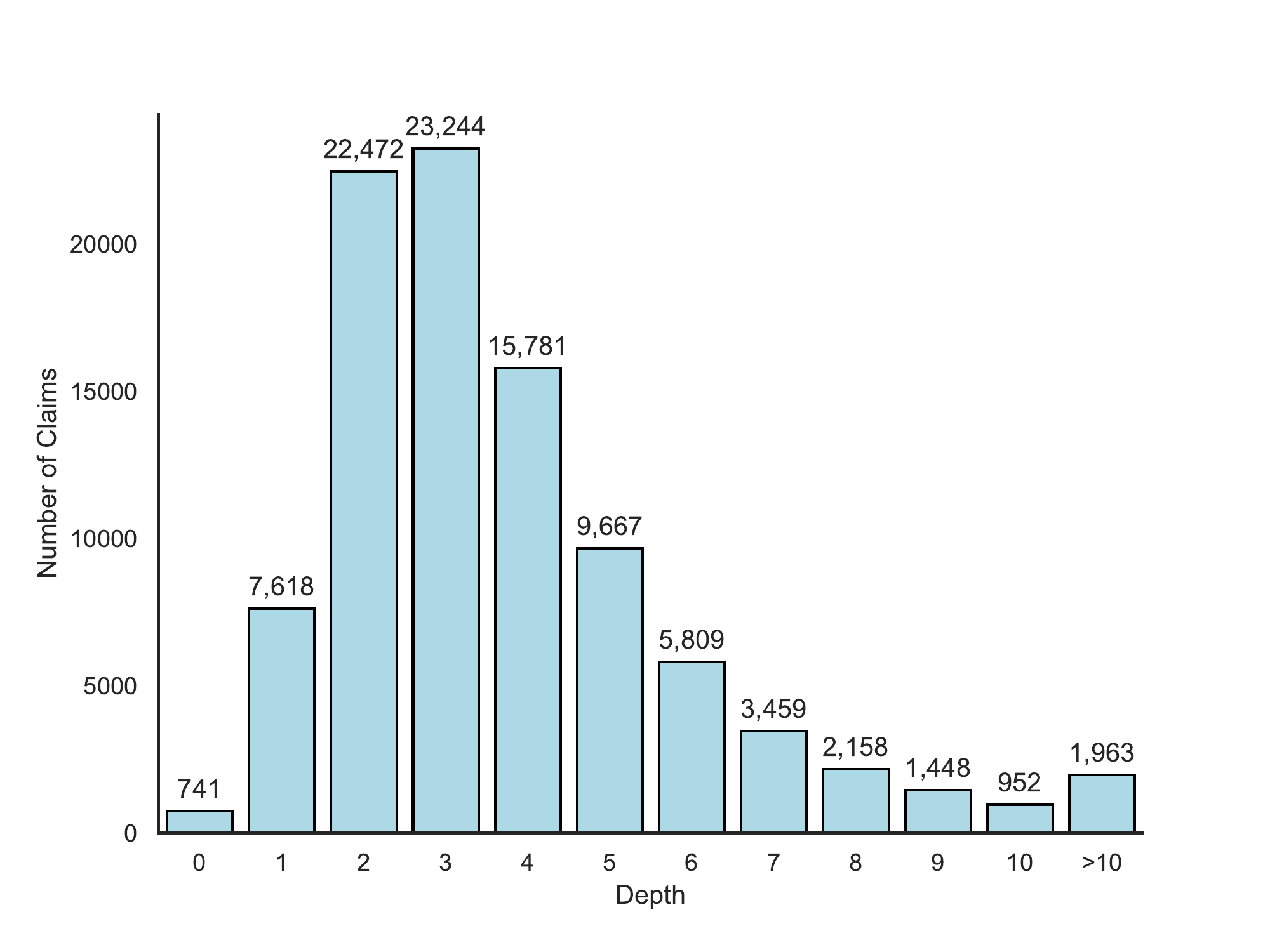}
\caption{Number of claims at given depths.}
\label{fig:num_node_level}
\end{figure}

Figure \ref{fig:num_node_level} shows the total number of claims at a given depth. We see that only $7,618$ out of $95,312$ claims directly support or oppose the theses of the controversial topics. The majority of the claims lie at depth $3$ or higher. This shows that the dataset has a rich set of supporting and opposing claims not only for the theses but for claims at different depths of the tree.

Moreover, around 47,000 claims in this dataset have \textbf{impact votes} assigned by the users of the platform. The impact vote evaluates how impactful a claim is within its context, which consists of its predecessor claims from the thesis of the tree. 
For example, claim \textbf{O1} {\sc``It is morally wrong to harm a defenseless person''} is an opposing claim for the thesis, and it is an {\sc impactful claim} since most of its impact votes belong to the category of {\sc very high impact}. However, claim \textbf{S3} {\sc``It is illegitimate for state actors to harm someone without the process''} is a supporting claim for its parent \textbf{O1} and it is a less impactful claim since most of the impact votes belong to the {\sc no impact} and {\sc low impact} categories. 

\begin{table}[h]
    \centering
    \begin{tabular}{|c|c|}
        \hline
       \# impact votes  & \# claims  \\
       \hline
       \hline
        $[3,5)$ & 4,495\\
        \hline
        $[5,10)$  & 5,405 \\
        \hline
        $[10,15)$  &  5,338\\
        \hline
        $[15,20)$ & 2,093\\
        \hline
        $[20,25)$ & 934\\
        \hline
        $[25,50)$ & 992\\
        \hline
        $[50,333)$ & 255\\
    
        \hline
    \end{tabular}
    \caption{Number of claims for the given range of number of votes. There are 19,512 claims in the dataset with $3$ or more votes. Out of the claims with $3$ or more votes, majority of them have $5$ or more votes.}
    \label{tab:vote_statistics_1}
\end{table}

\begin{table*}
    \centering
    \begin{tabular}{|c|c|c|}
        \hline
       & 3-class case & 5-class case \\
       \hline
       Agreement score  & Number of claims & Number of claims \\
       \hline
       \hline
       $>50\%$  & 10,848 & 7,304\\ 
       \hline
       $>60\%$ & 7,386 & 4,329\\ 
       \hline
       $>70\%$ & 4,412 & 2,195\\ 
       \hline
       $>80\%$ & 2,068 & 840\\ 
       \hline
   
    \end{tabular}
    \caption{Number of claims, with at least five votes, above the given threshold of agreement percentage for 3-class and 5-class cases. When we combine the low impact and high impact classes, there are more claims with high agreement score.}
    \label{tab:agreement_vote}
\end{table*}

\textbf{Impact label statistics.} Table \ref{tab:impact_label_stats} shows the distribution of the number of votes for each of the impact categories. The claims have $241,884$ total votes. The majority of the impact votes belong to {\sc medium impact} category. We observe that users assign more {\sc high impact} and {\sc very high impact} votes than {\sc low impact} and {\sc no impact} votes respectively. When we restrict the claims to the ones with at least $5$ impact votes, we have $213,277$ votes in total\footnote{26,998 of them {\sc no impact}, 33,789 of them {\sc low impact}, 55,616 of them {\sc medium impact}, 47,494 of them {\sc high impact} and 49,380 of them {\sc very high impact.}}.

\textbf{Agreement for the impact votes.} To determine the agreement in assigning the impact label for a particular claim, for each claim, we compute the percentage of the votes that are the same as the majority impact vote for that claim. 
Let $c_{i}$ denote the count of the claims with the class labels C=[{\sc no impact}, {\sc low impact}, {\sc medium impact}, {\sc high impact}, {\sc very high impact}] for the impact label $l$ at index $i$.

\begin{equation} \label{agreement}
\text{Agreement} = 100 * \frac{\max_{0 \leq i \leq 4}c_i}{\sum_{i=0}^{4} c_i}\%\\
\end{equation}

For example, for claim S1 in Figure \ref{fig:impact_image}, the agreement score is $100 * \frac{30}{90}\%=33.33\%$ since the majority class ({\sc no impact}) has $30$ votes and there are $90$ impact votes in total for this particular claim. We compute the agreement score for the cases where (1) we treat each impact label separately (5-class case) and (2) we combine the classes  {\sc high impact} and {\sc very high impact} into a one class: {\sc impactful} and {\sc no impact} and {\sc low impact} into a one class: {\sc not impactful} (3-class case). 

Table \ref{tab:agreement_vote} shows the number of claims with the given agreement score thresholds when we include the claims with at least $5$ votes. There are more claims with high agreement scores when we combine the low impact and high impact classes. This may imply that distinguishing between no impact-low impact and high impact-very high impact classes is difficult. In our experiments, we use a 3-class representation for the impact labels to decrease the sparsity issue. Moreover, to have a more reliable assignment of impact labels, we consider only the claims with have more than 60\% agreement. 

\textbf{Context.} In an argument tree, the claims from the thesis node (root) to each leaf node form an argument path. This argument path represents a particular line of reasoning for the given thesis. Similarly, for each claim, all the claims along the path from the thesis to the claim, represent the \textbf{context} for the claim. For example, in Figure \ref{fig:impact_image}, the context for \textbf{O1} consists of only the thesis, whereas the context for \textbf{S3} consists of both the thesis and \textbf{O1} since \textbf{S3} is provided to support the claim \textbf{O1} which is an opposing claim for the thesis.

\begin{table}[h]
    \centering
    \begin{tabular}{|c|c|}
        \hline
       Impact label  & \# votes- all claims \\
       \hline
       \hline
       No impact &  32,681\\ 
       \hline
       Low impact &  37,457\\ 
       \hline
       Medium impact &  60,136\\ 
       \hline
       High impact & 52,764\\ 
       \hline
       Very high impact & 58,846\\ 
       \hline
       Total \# votes & 241,884 \\ 
       \hline
    \end{tabular}
    \caption{Number of votes for the given impact label. There are $241,884$ total votes and majority of them belongs to the category {\sc medium impact}.  }
    \label{tab:impact_label_stats}
\end{table}
\textbf{Distribution of impact votes.} The distribution of claims with the given range of number of impact votes are shown in Table \ref{tab:vote_statistics_1}. There are 19,512 claims in total with $3$ or more votes. Out of the claims with $3$ or more votes, majority of them have $5$ or more votes.
We limit our study to the claims with at least $5$ votes to have a more reliable assignment for the accumulated impact label for each claim. 

\begin{table}
    \centering
    \begin{tabular}{|c|c|}
        \hline
       Context length  &  \# claims \\
       \hline
       \hline
       $1$ &  1,524 \\ 
       \hline
       $2$ & 1,977 \\ 
       \hline
       $3$ & 1,181\\ 
       \hline
       $[4,5]$ & 1,436 \\ 
       \hline
       $(5,10]$ & 1,115 \\ 
       \hline
       $>10$ & 153\\ 
       
       \hline
    \end{tabular}
    \caption{Number of claims for the given range of context length, for claims with more than $5$ votes and an agreement score greater than $60\%$.}
    \label{tab:context_length}
\end{table}

The claims are not constructed independently from their context since they are written in considering the line of reasoning so far. In most cases, each claim elaborates on the point made by its parent and presents cases to support or oppose the parent claim's points. Similarly, when users evaluate the impact of a claim, they consider if the claim is timely and appropriate given its context. 
There are cases in the dataset where the same claim has different impact labels when presented within a different context. Therefore, we claim that it is not sufficient to study only the linguistic characteristic of a claim to determine its impact, but it is also necessary to consider its context in determining the impact. 

\textit{Context length} ($\text{C}_{l}$) for a particular claim \textit{C} is defined by number of claims included in the argument path starting from the thesis until the claim \textit{C}. For example, in Figure \ref{fig:impact_image}, the context length for \textbf{O1} and \textbf{S3} are $1$ and $2$ respectively. Table \ref{tab:context_length} shows number of claims with the given range of context length for the claims with more than $5$ votes and $60\%$ agreement score. We observe that more than half of these claims have $3$ or higher context length. 

\section{Methodology}

\subsection{Hypothesis and Task Description}
Similar to prior work, we aim to understand the characteristics of impactful claims in argumentation. However, we \textbf{hypothesize} that the qualitative characteristics of arguments are not independent of the context in which they are presented. To understand the relationship between argument context and the impact of a claim, we aim to incorporate the context along with the claim itself in our predictive models.  

\textbf{Prediction task.} Given a claim, we want to predict the impact label that is assigned to it by the users: {\sc not impactful}, {\sc medium impact}, or {\sc impactful}.

\textbf{Preprocessing.} We restrict our study to claims with at least $5$ or more votes and greater than $60\%$ agreement to have a reliable impact label assignment. We have $7,386$ claims in the dataset satisfying these constraints\footnote{We have 1,633 {\sc not impactful}, 1,445 {\sc medium impact} and 4,308 {\sc impacful} claims.}. We see that the impact class {\sc impacful} is the majority class since around $58\%$ of the claims belong to this category.

For our experiments, we split our data to train (70\%), validation (15\%), and test (15\%) sets.

\subsection{Baseline Models}

\subsubsection{Majority} 
The majority baseline assigns the most common training example label ({\sc high impact}) to every test example.

\subsubsection{SVM with RBF kernel}  
Similar to \cite{habernal-gurevych-2016-makes}, we experiment with SVM with RBF kernel, with features that represent (1) the simple characteristics of the argument tree and (2) the linguistic characteristics of the claim. 

The features that represent the simple characteristics of the claim's argument tree include the distance and similarity of the claim to the thesis, the similarity of a claim with its parent, and the impact votes of the claim's parent claim. We encode the similarity of a claim to its parent and the thesis claim with the cosine similarity of their tf-idf vectors. The distance and similarity metrics aim to model whether claims which are more similar (i.e., \ potentially more topically relevant) to their parent claim or the thesis claim are more impactful. 

We encode the quality of the parent claim as the number of votes for each impact class and incorporate it as a feature to understand if it is more likely for a claim to be impactful given an impactful parent claim. 

\textbf{Linguistic features}. To represent each claim, we extracted the linguistic features proposed by \citet{habernal-gurevych-2016-makes} such as tf-idf scores for unigrams and bigrams, ratio of quotation marks, exclamation marks, modal verbs, stop words,  type-token ratio, hedging \citep{:/content/books/9789027282583}, named entity types, POS n-grams, sentiment \citep{ICWSM148109} and subjectivity scores \citep{wilson2005recognizing}, spell-checking, readibility features such as \textit{Coleman-Liau} \citep{1975-22007-00119750401}, \textit{Flesch} \citep{1949-01274-00119480601}, argument lexicon features \citep{somasundaran2007detecting} and surface features such as word lengths, sentence lengths, word types, and number of complex words\footnote{
We pick the parameters for the SVM model according to the performance validation split, and report the results on the test split.}.

\subsubsection{FastText}
\citet{joulin-etal-2017-bag} introduced a simple yet effective baseline for text classification, which they show to be competitive with deep learning classifiers in terms of accuracy. Their method represents a sequence of text as a bag of n-grams, and each n-gram is passed through a look-up table to get its dense vector representation. The overall sequence representation is simply an average over the dense representations of the bag of n-grams, and is fed into a linear classifier to predict the label. We use the code released by \citet{joulin-etal-2017-bag} to train a classifier for argument impact prediction, based on the claim text\footnote{We used maxNgram length of 2, learning rate of 0.8, num epochs of 15, vector dim of 300. We also used the pre-trained 300-dim wiki-news vectors made available on the fastText website.}.

\subsubsection{BiLSTM with Attention}
Another effective baseline \citep{zhou-etal-2016-attention, yang-etal-2016-hierarchical} for text classification consists of encoding the text sequence using a bidirectional Long Short Term Memory (LSTM) \citep{Hochreiter:1997:LSM:1246443.1246450}, to get the token representations in context, and then attending \citep{luong-etal-2015-effective} over the tokens to get the sequence representation. For the query vector for attention, we use a learned context vector, similar to \citet{yang-etal-2016-hierarchical}. We picked our hyperparameters based on performance on the validation set and report our results for the best set of hyperparameters\footnote{Our final hyperparams were: 100-dim word embedding, 100-dim context vector, 1 layer BiLSTM with 64 units, trained for 40 epochs with early stopping based on validation performance.}. We initialized our word embeddings with glove vectors \citep{pennington-etal-2014-glove} pre-trained on Wikipedia + Gigaword, and used the Adam optimizer \citep{DBLP:journals/corr/KingmaB14} with its default settings.

\begin{table*}[ht]
\centering
\begin{tabular}{|l|c|c|c|}
\hline
& Precision & Recall & F1  \\
\hline
Majority  & $19.43$ & $33.33$ & $24.55$\\
\hline 
\hline
SVM with RBF Kernel  &&&\\
\hline 
Distance from the thesis & $27.42$ & $33.53$ & $26.05$\\
\hline 
Parent quality  & $58.11$ & $47.85$ & $46.61$ \\
\hline
Linguistic features  & $\mathbf{65.67}$ & $38.58$ & $35.42$\\ 
\hline 
BiLSTM with Attention& $46.50_{\pm{0.28}}$ &$46.35_{\pm{0.99}}$& $46.22_{\pm{0.58}}$\\ 
\hline
FastText &  $51.18_{\pm{0.80}}$ &$46.09_{\pm{0.64}}$& $47.06_{\pm{0.70}}$\\ 
\hline
BERT models & & & \\
\hline
Claim only  & $53.24_{\pm{1.07}}$ &$50.93_{\pm{2.01}}$& $51.53_{\pm{1.53}}$\\ 
\hline 
Claim + Parent  & $55.79_{\pm{1.72}}$ &  $53.54_{\pm{2.09}}$ & $54.00_{\pm{1.79}}$\\ 
\hline
Claim + $\text{Context}_{f}(2)$  & $56.57_{\pm{0.85}}$ &  $54.76_{\pm{1.71}}$ & $55.18_{\pm{0.99}}$\\
\hline
Claim + $\text{Context}_{f}(3)$ &  $57.19_{\pm{0.92}}$ & $\mathbf{55.77_{\pm{1.05}}}$ & $\mathbf{55.98_{\pm{0.70}}}$\\
\hline 
Claim + $\text{Context}_{f}(4)$ &  $57.09_{\pm{1.71}}$ & $55.31_{\pm{1.09}}$ & $55.72_{\pm{1.14}}$\\
\hline
Claim + $\text{Context}_{gru}(4)$ &  $54.95_{\pm{2.00}}$ & $51.55_{\pm{1.27}}$ & $52.37_{\pm{1.26}}$\\
\hline
Claim + $\text{Context}_{a}(4)$   & $56.60_{\pm{0.52}}$ & $54.55_{\pm{0.57}}$ & $54.65_{\pm{0.33}}$ \\ 
\hline 

\end{tabular}
\caption{Results for the baselines and the BERT models with and without the context. Best performing model is BERT with the representation of previous $3$ claims in the path along with the claim representation itself. We run the models $5$ times and we report the mean and standard deviation. }
\label{tab:results}
\end{table*}

\subsection{Fine-tuned BERT model}
\citet{devlin2018bert} fine-tuned a pre-trained deep bi-directional transformer language model (which they call BERT) by adding a simple classification layer on top and achieved the state of the art results across a variety of NLP tasks. We employ their pre-trained language models for our task and compare them to our baseline models. For all the architectures described below, we fine-tune for 10 epochs, with a learning rate of 2e-5. We employ an early stopping procedure based on the model performance on a validation set.

\subsubsection{Claim with no context}
In this setting, we attempt to classify the impact of the claim based on the text of the claim only. We follow the fine-tuning procedure for sequence classification detailed in \citet{devlin2018bert}, and input the claim text as a sequence of tokens preceded by the special [CLS] token and followed by the special [SEP] token. We add a classification layer on top of the BERT encoder, to which we pass the representation of the [CLS] token and fine-tune this for argument impact prediction.

\subsubsection{Claim with parent representation} 
In this setting, we use the parent claim's text, in addition to the target claim text, in order to classify the impact of the target claim. We treat this as a sequence pair classification task and combine both the target claim and parent claim as a single sequence of tokens, separated by the special separator [SEP]. We then follow the same procedure above for fine-tuning.

\subsubsection{Incorporating larger context} 
In this setting, we consider incorporating a larger context from the discourse in order to assess the impact of a claim. In particular, we consider up to four previous claims in the discourse (for a total context length of 5). We attempt to incorporate larger context into the BERT model in three different ways.

\textbf{Flat representation of the path.}
The first, simple approach is to represent the entire path (claim + context) as a single sequence, where each of the claims is separated by the [SEP] token. BERT was trained on sequence pairs, and therefore the pre-trained encoders only have two segment embeddings \citep{devlin2018bert}. So to fit multiple sequences into this framework, we indicate all tokens of the target claim as belonging to segment A and the tokens for all the claims in the discourse context as belonging to segment B. This way of representing the input requires no additional changes to the architecture or retraining, and we can just fine-tune in a similar manner as above. We refer to this representation of the context as a flat representation, and denote the model as $\text{Context}_{f}(i)$, where $i$ indicates the length of the context that is incorporated into the model.

\begin{table*}[ht]
\centering
\begin{tabular}{|l|c|c|c|c|c|}
\hline
& $\text{C}_{l}=1$ & $\text{C}_{l}=2$ & $\text{C}_{l}=3$ & $\text{C}_{l}=4$ \\
\hline
BERT models & & & &   \\
\hline
\hline
Claim only & $48.61_{\pm{3.16}}$ & $53.15_{\pm{1.95}}$ & $54.51_{\pm{1.91}}$ &  $50.89_{\pm{2.95}}$\\ 
\hline 
Claim + Parent & $51.49_{\pm{2.63}}$ & $54.78_{\pm{2.95}}$& $54.94_{\pm{2.72}}$ & $51.94_{\pm{2.59}}$ \\
\hline
Claim +  $\text{Context}_{f}(2)$ & $52.84_{\pm{2.55}}$ & $53.77_{\pm{1.00}}$ & $55.24_{\pm{2.52}}$ &  $57.04_{\pm{1.19}}$\\ 
\hline
Claim + $\text{Context}_{f}(3)$  & $\mathbf{54.88_{\pm{2.49}}}$ &  $54.71_{\pm{1.74}}$ & $52.93_{\pm{2.07}}$& $\mathbf{58.17_{\pm{1.89}}}$ \\
\hline 
Claim + $\text{Context}_{f}(4)$ &  $54.47_{\pm{2.95}}$ & $\mathbf{54.88_{\pm{1.53}}}$ &$\mathbf{57.11_{\pm{3.38}}}$ & $57.02_{\pm{2.22}}$ \\
\hline
\end{tabular}
\caption{F1 scores of each model for the claims with various context length values.}
\label{tab:results_context}
\end{table*}

\textbf{Attention over context.}
Recent work in incorporating argument sequence in predicting persuasiveness \citep{DBLP:conf/aaai/HideyM18} has shown that hierarchical representations are effective in representing context. Similarly, we consider hierarchical representations for representing the discourse. 
We first encode each claim using the pre-trained BERT model as the claim encoder and use the representation of the [CLS] token as claim representation. 
We then employ dot-product attention \citep{luong-etal-2015-effective}, to get a weighted representation for the context. We use a learned context vector as the query for computing attention scores, similar to \citet{yang-etal-2016-hierarchical}. The attention score $\alpha_c$ is computed as shown below:
\begin{equation}
    \alpha_{c} = \frac{exp(V_{c}^T V_{l})}{\sum_{c\in{D}} exp(V_{c}^T V_{l})}
\end{equation}

Where $V_c$ is the claim representation that was computed with the BERT encoder as described above, $V_l$ is the learned context vector that is used for computing attention scores, and $D$ is the set of claims in the discourse.
After computing the attention scores, the final context representation $v_d$ is computed as follows:
\begin{equation}
     V_{d} = \sum_{c\in{D}}\alpha_{c} V_{c}
\end{equation}
We then concatenate the context representation with the target claim representation $[V_d, V_r]$ and pass it to the classification layer to predict the quality. We denote this model as $\text{Context}_{a}(i)$.

\textbf{GRU to encode context}
Similar to the approach above, we consider a hierarchical representation for representing the context. We compute the claim representations, as detailed above, and we then feed the discourse claims' representations (in sequence) into a bidirectional Gated Recurrent Unit (GRU) \citep{cho-al-emnlp14}, to compute the context representation. We concatenate this with the target claim representation and use this to predict the claim impact. We denote this model as $\text{Context}_{gru}(i)$.

\section{Results and Analysis}
Table \ref{tab:results} shows the macro precision, recall, and F1 scores for the baselines as well as the BERT models with and without context representations\footnote{For the models that result in different scores with a different random seed, we run them $5$ times and report the mean and standard deviation.}. 

We see that \textit{parent quality} is a simple yet effective feature, and the SVM model with this feature can achieve significantly higher ($p<0.001$)\footnote{We perform a two-sided t-test for significance analysis.} F1 score ($46.61\%$) than \textit{distance from the thesis} and \textit{linguistic features}. Claims with higher impact parents are more likely to have a higher impact. \textit{Similarity with the parent and thesis} is not significantly better than the \textit{majority} baseline. Although the BiLSTM model with attention and FastText baselines performs better than the SVM with \textit{distance from the thesis} and \textit{linguistic features}, it has similar performance to the \textit{parent quality} baseline.

We find that the BERT model with \textit{claim only} representation performs significantly better ($p<0.001$) than the baseline models.  Incorporating the \textit{parent representation} only along with the \textit{claim representation} does not give significant improvement over representing the claim only. However, \textit{incorporating the flat representation of the larger context} along with the claim representation consistently achieves significantly better ($p<0.001$) performance than the claim representation alone. Similarly, \textit{attention representation} over the context with the learned query vector achieves significantly better performance then the \textit{claim representation} only ($p<0.05$). 

We find that the \textit{flat representation} of the context achieves the highest F1 score. It may be more difficult for the models with a larger number of parameters to perform better than the \textit{flat representation} since the dataset is small. We also observe that modeling $3$ claims on the argument path before the target claim achieves the best F1 score ($55.98\%$).

To understand for what kinds of claims the best performing contextual model is more effective, we evaluate the BERT model with \textit{flat context representation} for claims with context length values $1$, $2$, $3$ and $4$ separately. Table \ref{tab:results_context} shows the F1 score of the BERT model without context and with \textit{flat context representation} with different lengths of context. For the claims with context length $1$, adding $\text{Context}_{f}(3)$ and $\text{Context}_{f}(4)$ representation along with the claim achieves significantly better $(p<0.05)$ F1 score than modeling the \textit{claim only}. Similarly for the claims with context length $3$ and $4$, $\text{Context}_{f}(4)$ and $\text{Context}_{f}(3)$ perform significantly better than BERT with \textit{claim only} ($(p<0.05)$ and $(p<0.01)$ respectively). We see that models with larger context are helpful even for claims which have limited context (e.g., $\text{C}_{l}=1$). This may suggest that when we train the models with larger context, they learn how to represent the claims and their context better. 

\section{Limitations}
In this study, we find that incorporating pragmatic context is crucial in impact prediction. First, we present a new dataset for this task. We assume that the impact labels in this dataset are provided in good faith by the users. However, we note that the user demographics on the platform may not have a fair representation, and prior beliefs and background could affect which arguments are perceived as more impactful. We should account for this potential bias while using the systems built from this dataset. We further observe that BERT-based models achieve the best predictive performance. However, it is difficult to interpret these systems to understand what aspect of the context plays an important role. In future work, we aim to employ methods such as local surrogate \citep{10.1145/2939672.2939778} or input saliency models \citep{li-etal-2016-visualizing} to interpret these systems.

\section{Chapter Summary}

This chapter proposes a new dataset of arguments along with their impact label and the argument path, representing a particular line of reasoning on the given controversial topic. We further propose predictive models that incorporate the pragmatic and discourse context of argumentative claims to predict argument impact. We show that the models representing the pragmatic context outperform models that rely on only claim-specific linguistic features for predicting the perceived impact of individual claims within a particular line of argument.

\chapter{Conclusion and Future Work}\label{conclusion}

In this dissertation, we describe our contributions to understand persuasion in computational argumentation. In particular, we show that the characteristics of the people involved highly influence the process of persuasion.  
We investigate the impact of speaker and audience factors in predicting the more persuasive debater. 
We also explore whether a user's social interaction on online argumentation platforms affects their success in persuasion over time. We further propose context-aware models to measure the importance of pragmatic context in predicting the impact of the arguments. 

\section{Summary of Contributions}

In Chapter \ref{dataset}, we propose a new dataset of debates with extensive user information extracted from an online argumentation platform (i.e., \href{debate.org}{\textit{debate.org}}). This is the largest available dataset with such extensive user information, including political ideology, religious ideology, and stance on various controversial topics. Availability of this information has motivated further research in exploring the effect of user factors in persuasion \citep{10.1162/tacl_a_00281,Durmus:2019:MFU:3308558.3313676}. 

With the dataset in hand, we study the role of prior beliefs, of both speakers and audience members, on the perceived persuasiveness of arguments.
We do this by formulating a new task to determine which debater will be able to persuade a given voter to change their stance.
We find that features associated with a user's initial stance are very predictive for this task.
This is especially true for debates on political and religious issues, where these features are even more predictive than linguistic features of the arguments.

In Chapter \ref{social_interactions}, we further explore whether a user's social interaction impacts their debating success over time on online argumentation platforms.
We extract features from a user's friendship and voter networks. 
We then use these features to explore the role of social interactions as compared to personality traits and language in predicting debating success over time.
We find that social interaction features (i.e., primarily features extracted from the voter network) are the most predictive of success. We observe that the best predictive performance is achieved when combining social interaction features with 
linguistic features.
This implies that the characteristics of interactions on online debating platforms are essential to becoming more experienced and successful in persuasion. 

Finally, we propose a dataset to study the role of \textit{kairos} (i.e., pragmatic context) in determining argument impact. As described in Chapter \ref{context}, the dataset includes the argument context for each claim, along with the impact score within the given line of reasoning. We further explore whether a flat vs. a hierarchical representation of context is more effective for this task. We find that a flat representation of the context achieves the best performance since the dataset may not be large enough to learn the additional parameters needed for a hierarchical model. We observe that models that incorporate context perform significantly better than those that use the claim only. 
This implies that the context in which an argument is presented is crucial in assessing its impact.

\section{Ethical Considerations}

All the data in our research has been collected and used in accordance with the terms of service of the source. For user studies, we take the utmost care in making sure that the anonymity of the users is preserved. Finally, we make sure that our work does not take a stance on any of the controversial topics, but rather just analyzes the viewpoints of the participants in the datasets we use. One shortcoming we acknowledge is that we are unable to represent all demographics due to a lack of data. The sources we used tend to be highly skewed towards an American audience, and even within this audience, the distribution may not be representative enough. 

Given that argumentation is a fundamental part of human communication, the work in this area could be used in both good and ethically less acceptable manners. The driving motivator of this dissertation has always been that argumentation can be used for social good, such as exposing people to diverse viewpoints to help them make more informed decisions or using persuasion to encourage people to contribute to the environment and society. However, even for such use cases, it is vital to be transparent and inform users about the nature of these systems. Moreover, user consent should be required to employ such methods in real-world scenarios. 

\section{Future Directions}

 \textbf{Modeling Users in Computational Persuasion.} In our study, we explore the role of prior beliefs in persuasion, focusing on political and religious ideologies. However, there are many aspects of the source and the audience (e.g., education level, prior argumentation skills, credibility, personality traits) that may influence the persuasion process. It is challenging to control for all potential confounding factors to isolate the effect of the linguistic features due to data sparsity. We think it is vital to explore better representations for users to disentangle the impact of user aspects. We further want to explore the following research questions: 1) How do different aspects of users influence their perceptions of the arguments? 2) How do these aspects affect people's language choice while interacting with more similar vs. different people? 3) How does the language use change for different groups of speakers?

\textbf{Personalized Argument Generation.} Understanding the effect of user factors in persuasion could be the first step towards designing personalized argument generation systems capable of conveying relevant and interesting information for a more effective persuasion process \citep{edseee.928406920201001, dijkstra_2000}. Personalization is important in increasing engagement and attachment in social interactions on online platforms \citep{edselc.2-52.0-8497816854120160801, edsemr.10.1108.AJIM.03.2018.006720180907}. Therefore, having personalized systems may increase the quality of persuasive communication and the outcome of this process. For example, \cite{wang-etal-2019-persuasion} has recently proposed a personalized dialogue system that tries to persuade people to donate to a specific charity. They show that personalized argumentation generation systems can be used for social good. Moreover, such systems could be used to present people with a diverse set of viewpoints to help them make more informed decisions. 

\textbf{Interpretation of Neural Models.} Neural networks can model more complex representations that help achieve state-of-the-art performance in various syntactic and semantic tasks in Natural Language Processing. However, unlike feature-based linear models, it is more challenging to interpret neural models to explain what these models learn and improve them. For example, in Chapter 6, we have found that incorporating context with the argument helps predict its impact. However, it is not straightforward to interpret explicitly which aspects of the context helps to improve the overall performance. Similarly, although neural methods achieve state-of-the-art performance in persuasion prediction tasks, it is difficult to identify the characteristics of persuasive language and effective persuasion strategies. We believe that improved interpretation of neural networks is crucial to draw valuable conclusions in computational persuasion studies and build better models for these tasks.

\bibliography{durmus_thesis}
\bibliographystyle{thesis_natbib}

\end{document}